\documentclass[conference]{IEEEtran}
\IEEEoverridecommandlockouts
\usepackage{cite}
\usepackage{amsmath,amssymb,amsfonts}
\usepackage{algorithmic}
\usepackage{graphicx}
\usepackage{textcomp}
\usepackage{booktabs} 
\usepackage[ruled,vlined]{algorithm2e}
\usepackage{subfig}
\usepackage{float}
\usepackage{graphicx, float}
\usepackage{caption,lipsum}
\usepackage{makecell}

\usepackage{tabulary}
\usepackage{multirow}
\usepackage{array}

\newcommand\Tstrut{\rule{0pt}{2.6ex}}         

\def\BibTeX{{\rm B\kern-.05em{\sc i\kern-.025em b}\kern-.08em
    T\kern-.1667em\lower.7ex\hbox{E}\kern-.125emX}}
    
\usepackage{fancyhdr} 
\usepackage{kantlipsum} 
\fancypagestyle{plain}{ 
\fancyhf{} 
\fancyhead[C]{Conference on \LaTeX} 
 
} 
\usepackage{eso-pic} 
\begin{document} 
    

\title{Shortest Path Distance Approximation using Deep learning Techniques}


%

\author{\IEEEauthorblockN{Fatemeh Salehi Rizi}
\IEEEauthorblockA{\textit{Department of Computer Science}\\
\textit{and Mathematics}\\
\textit{University of Passau}\\
Passau, Germany \\
Fatemeh.SalehiRizi@uni-passau.de}
\and
\IEEEauthorblockN{Joerg Schloetterer}
\IEEEauthorblockA{\textit{Department of Computer Science } \\
\textit{and Mathematics}\\
\textit{University of Passau}\\
 Passau, Germany\\
Joerg.Schloetterer@uni-passau.de}
\and
\IEEEauthorblockN{Michael Granitzer}
\IEEEauthorblockA{\textit{Department of Computer Science} \\
\textit{and Mathematics}\\
\textit{University of Passau}\\
Passau, Germany \\
Michael.Granitzer@uni-passau.de}
}

\maketitle

\begin{abstract}

Computing shortest path distances
between nodes lies at the heart of many graph algorithms and applications.
Traditional exact methods such as breadth-first-search (BFS) do not scale
up to contemporary, rapidly evolving today's massive
networks. Therefore, it is required to find approximation methods to enable scalable graph processing with
a significant speedup.
In this paper, we utilize vector embeddings learnt by
deep learning techniques to approximate the shortest
paths distances in large graphs. We show that a feedforward neural network fed with embeddings can approximate distances with relatively low distortion error.
The suggested method is evaluated on the Facebook,
BlogCatalog, Youtube and Flickr social networks.

\end{abstract}


\section{Introduction}
Finding shortest path distances between nodes in a graph is an
important primitive in a variety of applications.
For instance, the number of links
between two URLs indicates page similarity in a graph of the Web \cite{ref1}. In a semantic web ontology, shortest path distances among entities are used for ranking their relationships \cite{ref2}.
The number of hops from one person to another indicates the level of trust in a trust network \cite{ref3}. In social networks, the shortest path distance is used to compute the closeness centrality \cite{ref4}.

Traditional methods for computing node distance do not
scale with graph size. For a graph with $n$ nodes and
$m$ edges, efficient implementations of Dijkstra compute the shortest paths for a node
to others in $O(n \log n + m)$ time. A slight generalization of Dijkstra, known as the $A^*$ algorithm,
uses heuristic techniques for computing shortest distances. In practice, $A^*$ works at least as quickly as Dijkstra's algorithm, however, the run time complexity
is still $O(n \log n + m)$.
Tolerable for small graphs, but on a large million node graph computation can take up to a minute for a single node distance \cite{ref7}.
Given the high
cost of storing precomputed distances, researchers have
limited choice but to sample subgraphs or seek
approximate results. For many practical applications, finding out approximate
distances between nodes can be sufficient.

In this paper, we propose a novel method for approximating shortest path distance measurements between two nodes utilizing vector embeddings generated by deep learning techniques. Node2vec and  Poincar\`e embeddings are recently studied for the basic graph analysis tasks such as link prediction ~\cite{Grover2016, Nickel2017} and node classification~\cite{Grover2016,salehi2017}. This is due to the fact that they speed up the computation time and yield precise results compared to the traditional matrix factorization techniques. We thus exploit these two embeddings to approximate shortest path distances.
At a high level, we first learn embeddings using node2vec \cite{ref10} and Poincar\'{e} \cite{ref11} for every single node in the graph.
We then follow the "landmark-based" approach proposed in \cite{ref8}, where
we choose a small number of nodes as landmarks.
We compute the actual shortest paths distances from each landmark to all of the remaining nodes. 
We use these pairs as our training set. Finally, we train a feedforward neural network with embedding vectors to approximate the distance between corresponding nodes. Feedforward networks are well known for their good representational capability \cite{bilski2005ud} and should be sufficient to approximate shortest path distance functions giving graph embeddings.


Due to the sparsity property of real-world (social) networks, the number of edges is proportional to the number of nodes. Therefore, generating the ground truth using the landmark-based approach takes a linear time complexity which yields a linear running time for the entire proposed method. More details are elaborated in section \ref{sec:app}.

Overall, we study advanced graph embedding techniques to make the following contributions:
\begin{itemize}
\item We theoretically motivate and suggest the use of node2vec and Poincar\'{e} embeddings for node distance approximation in large graphs (section \ref{sec:app}).


\item We show that neural networks can predict the shortest path distances effectively and efficiently, especially for shorter paths (section \ref{sec:exp}).

\item We demonstrate that different embedding techniques as well as different parameter settings have a significant influence on the approximation quality (section \ref{sec:exp}).

\item We compare our approximations to the most prominent works in the state-of-the-art and show, that our approach outperforms current methods in terms of prediction accuracy (section \ref{sec:exp}).

\end{itemize}

In order to demonstrate the performance of our method in practice, we conduct experiments
with four real-world datasets. The experiments indicate, that our predictor performs better for shorter paths. However, embeddings behave poorly to approximate the longer paths. The reason lies in both the embedding technique and the sampling strategy for training pairs. We provide more
details in section \ref{sec:exp}.

The rest of the paper is structured as follows. Section \ref{sec:background} provides an overview of recent existing  shortest path distance approximation and graph embedding techniques. Our proposed method is elaborated in section \ref{sec:app}. Section \ref{sec:exp} summarizes the
experimental evaluation results. Finally, section \ref{sec:con} concluding remarks.

\section{Background and related work}
\label{sec:background}
\subsection{Shortest Path}
\label{ssec:shortes-path}

The task of computing the shortest path distance from a single node to all other nodes is known as
\textit{single source shortest paths (SSSP)}. 
Exact methods such as Dijkstra compute SSSP for
weighted graphs with $n$ nodes and $m$ edges in time $O(m+n \log n)$. For unweighted sparse graphs,
shortest paths can be computed using Breadth First Search
(BFS) in time $ O(m+n)$.
However, exact methods, are extremely slow for performing queries on today's very large online networks. 
For many practical applications, finding out approximate
distances between nodes can be sufficient. Among the approximate methods, a family of scalable algorithms for this problem are the so-called landmark-based approaches \cite{ref8}. In this family of
techniques, a fixed set of landmark nodes is selected and actual shortest paths are precomputed from each
landmark to all other nodes. Knowledge of the distances to the landmarks, together with the triangle inequality, typically allows one to compute approximate distance between any two nodes in $O(l)$ time, where $l$ is the number of landmarks. \cite{ref8}. Although landmark-based algorithms do not provide strong
theoretical guarantees on approximation quality \cite{ref13}, they
have been shown to perform well in practice, scaling up to
graphs with millions of edges with acceptable accuracy and response times of under one second per
query \cite{ref14}.


Inspired by landmark-based methods, authors in \cite{ref8} and \cite{ref9}, suggested the idea of graph coordinate systems, which embeds
graph nodes into points on a coordinate space. The resulting
coordinates can be used to quickly approximate node distance
queries on the original graph. 
However, it has several limitations in practice. First, their initial
graph embedding process is centralized and computationally
expensive, which presents a significant performance bottleneck
for larger graphs. Second, their results produce relatively high error rates, which limits the types of applications
it can serve. Finally, it is unable to produce actual paths
connecting node pairs, which is often necessary for a number
of graph applications.

\subsection{Graph Embedding}
\label{ssec:graph-embedding}
Besides the aforementioned methods, embedding the nodes with the explicit objective of preserving the shortest path length, various methods to embed the graph in general (or dimension reduction respectively) have been proposed (c.f. Goyal and Ferrara~\cite{goyal2017} for a survey). Among the classical methods are Principal Component Analysis (PCA)~\cite{jolliffe1986}, Linear Discriminant Analysis (LDA)~\cite{martinez2001}, ISOMAP~\cite{tenenbaum2000}, Multidimensional Scaling (MDS)~\cite{kruskal1978}, LLE~\cite{roweis2000} and Laplacian Eigenmap~\cite{belkin2002} (c.f. Yan et al.~\cite{yan2007} for a survey). However, most of these methods typically rely on solving eigen decomposition and the complexity is at least quadratic in the number of nodes, which makes them inefficient to handle large-scale networks.

Recently, neural network based approaches have been proposed, which were inspired by the ideas of Word2Vec~\cite{mikolov2013}, which is build around the distributional hypothesis, stating that words in similar contexts tend to have similar meaning~\cite{harris54}. DeepWalk~\cite{perozzi2014} samples random walks from the graph and treats them as sentence equivalents. That is, given the representation of a node in the embedding space, DeepWalk approximates the conditional probability of nodes in the the neighborhood. Thereby, nodes sharing a similar neighborhood, tend to have a similar representation in the embedding space. Similar to Word2Vec implicitly factorizing a matrix of word co-occurrences~\cite{pennington2014,Levy2014}, Deepwalk has been shown to factorize a matrix of node transition probabilities~\cite{Yang2015}. Node2vec~\cite{Grover2016} extends Deepwalk by introducing parameters to control the random walk behaviour. At the most extreme parameter choices, node2vec employs breadth-first or depth-first sampling, exploring the close-by neighborhood or nodes that are far apart in the network. LINE~\cite{Tang2015} explicitly optimizes the embeddings to capture first- and second-order proximity, by training separate embeddings for them, which are finally concatenated. First-order proximity is given by explicit connections between nodes, while second-order proximity is given by comparing the nodes' neighborhoods. 
Nickel and Kiela~\cite{Nickel2017} embed the graph into a hyperbolic space, or more precisely into an n-dimensional Poincar\'{e} ball, capturing hierarchy and similarity. Several approaches have been proposed to better model long distance relationships or higher order proximity respectively. Instead of approximating the k-order proximity matrix, as DeepWalk does, GraRep~\cite{cao2015} calculates it accurately, at the cost of increased complexity. Yang et al.~\cite{yang2017} alleviate this problem by using information from lower order proximity matrices. The authors of HOPE~\cite{ou2016} experimented with different similarity measures, such as Katz Index, Rooted Page Rank and Adamic-Adar.
HARP~\cite{chen2018} and Walklets~\cite{perozzi2017} address capturing higher-order proximity by adapting the random walk strategy. While HARP coarsens the graph and learns representations via hierarchically collapsed graphs, Walklets skips over steps in the random walks. 
Deep architectures have been proposed, aiming at capturing non-linearity in the graphs. SDNE~\cite{wang2016} and DNGR~\cite{cao2016} utilize autoencoders, GCN~\cite{kipf2017} defines a convolution operator on the graph. 

 To our best knowledge, approximating the shortest path distances based on embeddings that have not been specifically tailored towards this end, has not been investigated yet.

\section{Approach}
\label{sec:app}

\subsection{Distance Approximation}
\label{ssec:distance}
Let $G = (V, E)$ be an unweighted undirected graph with $n$ nodes and $m$ edges. Graph embedding techniques
create a real-valued, the so called vector embedding  $\phi (v) \in R^d$ for every node $v \in V$.
Given a pair of nodes $u,v \in V$ with the real shortest path distance $d_{u,v}$, the goal is to approximate the distance as $\hat{d}$ using a feedforward neural network. Formally, we define $\hat{d}$ as function
\[ \hat{d}:  \phi(u) \times \phi(v)  \mapsto R^+ \]
that maps a pair of vector embeddings to a real-valued shortest path distance $d_{u,v}$ in $G$.

To train the neural network, we need to extract training pairs from the entire graph $G$.
We first choose a small number of $l$ nodes as landmarks,
where $l \ll n$. We then compute the actual shortest distances from each landmark to all of the remaining nodes using BFS. 
It yields $ l(n-l)$ training pairs. Given a training pair $<\phi(v),\phi(u)>$, we create a joint representation as input to the neural network by applying a binary operation, namely subtraction, concatenation, average and point-wise multiplication, over the vector embeddings. The definitions of the binary operations are listed in the Table~\ref{tab:opr}. Eventually, vectors of the training set serve as input for a feedforward neural network. The neural network maps the input vectors to a real-valued distance.

\begin{table}[htbp]
\caption{Choice of binary operators correspond to the $ith$ component of $\phi$}
\begin{center}
\begin{tabular}{|c|c|c|}
\hline
\textbf{Operator}&\textbf{Symbol} &\textbf{Definition} \Tstrut  \\ 
\hline
Subtraction & $\ominus $ & $\phi_i(u)-\phi_i(v)$ \Tstrut \\
\hline
 Concatenation & $\oplus$&  $(\phi(u),\phi(v))$ \Tstrut  \\
\hline
Average  &  $\oslash$ & $\frac{\phi_i(u)+\phi_i(v)}{2}$ \Tstrut  \\
\hline
Hadamard & $ \odot$  &$\phi_i(u)*\phi_i(v)$ \Tstrut \\
\hline

\end{tabular}
\label{tab:opr}
\end{center}
\end{table}

Our feedforward network consists of an input layer, a hidden layer,
and an output layer. The size of the input layer depends on the binary operation on vector embeddings. For example subtraction needs $d$ neurons while concatenation requires $2d$.
We set the rectified unit (ReLU)~\cite{nair2010rectified} as activation function for the first two layers.
ReLU does not face gradient vanishing problem and it has been shown that deep networks are trained efficiently using ReLU. Since the network does a regression task, the output layer is a single unit of softplus~\cite{glorot2011deep} which is a smoother version of ReLU with the range of $[0,\infty]$. 
We assess the quality of predictor by Mean Squared Error (MSE) which measures the average of the squares of difference between the estimator and what is estimated. As optimizer, we use Stochastic Gradient Descent (SGD)~\cite{bottou2010large} which is usually fast and efficient for large-scale learning.

\subsection{Computational Complexity}
\label{ssec:comp}
The proposed method achieves a linear runtime complexity. First, we learn vector embeddings which takes precomputation time $O(n)$ where $n $ is the number of nodes in the graph ~\cite{goyal2017}.

We then use the landmark-based scheme to minimize the number of shortest path computations
needed to establish the ground truth. By choosing a
small, constant number of landmarks, we only need to compute
a BFS tree for each landmark. The resulting values represent
shortest distances from all remaining nodes to these
landmarks, and are sufficient to compose the training set. With $l$ nodes as landmarks,
where $l \ll n$,  we have $ l (n-l)$ training pairs. It takes time $O(l(n+m))$, knowing BFS on unweighted sparse graphs consumes $O(n+m)$ time.
The advantage of using a graph embedding is that a feedforward neural network can answer a
distance query between two nodes $u$, $v$ in a small amount of time independent of the
graph size, i.e. $O(1)$ time. So calculating the shortest path distances from a starting node $u$ to all other nodes takes $O(n)$.


\section{Experimental Evaluation}
\label{sec:exp}

In this section, we report the performance of the proposed method. We evaluate node2vec and Poincar\'{e} embeddings for distance approximation applying the four different binary operations.
We start by describing
our datasets, and then we set the hyperparameters.
Eventually, we describe our verification approach along with discussions over the results.

\subsection{Datasets}
\label{subsec:dataset}
We test our method on four real-world social network
graphs, representing four different orders of magnitude in
terms of network size.
\begin{itemize}


\item  \textbf{Facebook.}
This dataset consists of friends lists from Facebook. Facebook data was collected from survey participants using a Facebook app. Facebook data has been anonymized by replacing the Facebook internal ids for each user with a new value~\cite{leskovec2012learning}. 

\item  \textbf{BlogCatalog.} 
This is a network of social relationships
of the bloggers listed on the BlogCatalog website. The labels
represent blogger interests inferred through the metadata
provided by the bloggers~\cite{Zafarani+Liu:2009}.

\item  \textbf{Youtube.} This is the friendship network of the video-sharing site Youtube. Nodes are users and an undirected edge between two nodes indicates a friendship~\cite{yang2015defining}.

\item  \textbf{Flickr.} This dataset is built by forming links between images sharing common metadata from Flickr. Edges are formed between images from the same location, submitted to the same gallery, group, or set, images sharing common tags, images taken by friends~\cite{mislove2007measurement}.

\end{itemize}

The properties of the datasets are summarized in Table~\ref{tab:dataset}. The table shows the number of nodes $n$, number
of edges $m$ and average shortest distance between nodes $\overline{d}$. The most relevant previous works using a similar methodology, Orion~\cite{ref8} and Rigel~\cite{ref9}, used Flickr as a common dataset in their evaluations. We therefore compare our results with the state-of-the-art on the Flickr dataset.

\begin{table}[htbp]
\caption{Statistics of Social Network Datasets}

  \label{tab:dataset}
 
 \begin{center}
\begin{tabular}{|l |c|c|c|}

 \hline
  \textbf{Dataset } &   \textbf{Nodes } & \textbf{Edges } & $\overline{d}$ \Tstrut \\
 
  \hline
  
  Facebook    & 4,039 & 88,234 & 4.31 \Tstrut \\
  
  \hline
  BlogCatalog    & 10,312 & 333,983 & 2.72 \Tstrut  \\
  
  \hline
  Youtube    & 1,134,890 & 2,987,624 & 5.55 \Tstrut \\
  
  \hline
  Flickr  & 1,715,255 & 15,551,250 & 5.13 \Tstrut \\
  
 \hline
\end{tabular}
\end{center}

\end{table}

\subsection{Parameters and Environment}
\label{subsec:param}

To generate node2vec embeddings, we use the default settings ~\cite{Grover2016} with the number of walks $\gamma=10$ and length $l=80$. We tuned the context size setting it finally to $k=5$ for learning node representations by
predicting more nearby nodes. We also set $p=q=1$ to explore local and global nodes equally.
For Poincar\'{e}, we follow the default settings as $r>1$, $t=0.01$ \cite{Nickel2017}, and $50$ iterations. 

We consider two different embedding dimensions $d = 32$ and $d = 128$ to investigate the impact of more features on distance approximation. To train the neural network, we initialize weights randomly and set the learning rate $lr=0.01$  running $15$ iterations. We run experiments on Linux machines with
five 3.50GHz Intel Xeon(R) CPUs and 16GB memory.

\subsection{Approximation Quality}
\label{subsec:err}

Since shortest path distances are discrete real values, we first round the results of the prediction.
We then measure the accuracy of our method using two key metrics. The first is the Mean of Relative Error (MRE).
The Relative Error is widely used in the study of shortest path evaluation and defined as,

\[ RE= \frac{|\hat{d}-d|}{d} \]
where $d$ is the actual distance measured by BFS algorithm and $\hat{d}$ the approximation \cite{ref8, ref13, ref14}. The MRE has a natural tendency to be smaller for larger distance values, which is not a perfectly fair comparison. Therefore, as a second evaluation measure we use the Mean Absolute Error (MAE) which is the most natural measure of average error magnitude independent of the target value. 
%

\subsection{Training and Test Data}
\label{fig:train}

In order to capture training pairs, we  require computing BFS trees rooted
from each landmark to all other remaining nodes.
For the smaller datasets such as Facebook and BlogCatalog, we set $l=100$ while for the two others $5$ landmarks are enough.
We omit paths with length $1$ since the time complexity of finding direct neighbors is already linear $O(n)$ \cite{kang2011centralities}.

To avoid an imbalanced training set and perform a fast sampling, we downsample the minority classes (i.e. classes with few samples). In detail, when we search the shortest path for a given training pair by BFS, we also retrieve the nodes are included in this path. Respecting the fact that a shortest path carries nodes by keeping shortest distance between them. Additionally, in complex networks such as social networks the average distance is very small therefore the distribution of long paths is quite low~\cite{akiba2013fast}. In Figure~\ref{fig:train}, we plot
histograms that reflect the distribution of training pairs in all four datasets. 
In BlogCatalog the average shortest path distance is $2.72$ and the distribution of path longer than $5$ is prohibitively low. We therefore limit the path length to $5$.

To compose the training matrix, we need to perform binary operations on the embedding vectors of given training pairs. We do Subtraction, Concatenation, Average and Hadamard on pairs of vectors. 


For test pairs, we do the same strategy as training pairs considering a smaller set of landmarks.
We start BFS traversals from landmarks to other remaining nodes which generate a set of unseen pairs.
Overall, we gather around $100,000$ unseen test pairs for each dataset. 
The statistics of training and test pairs are available in Table~\ref{tab:train}.

 \begin{figure*}[htbp]
\centering

\subfloat{
\includegraphics[width=.24\textwidth]{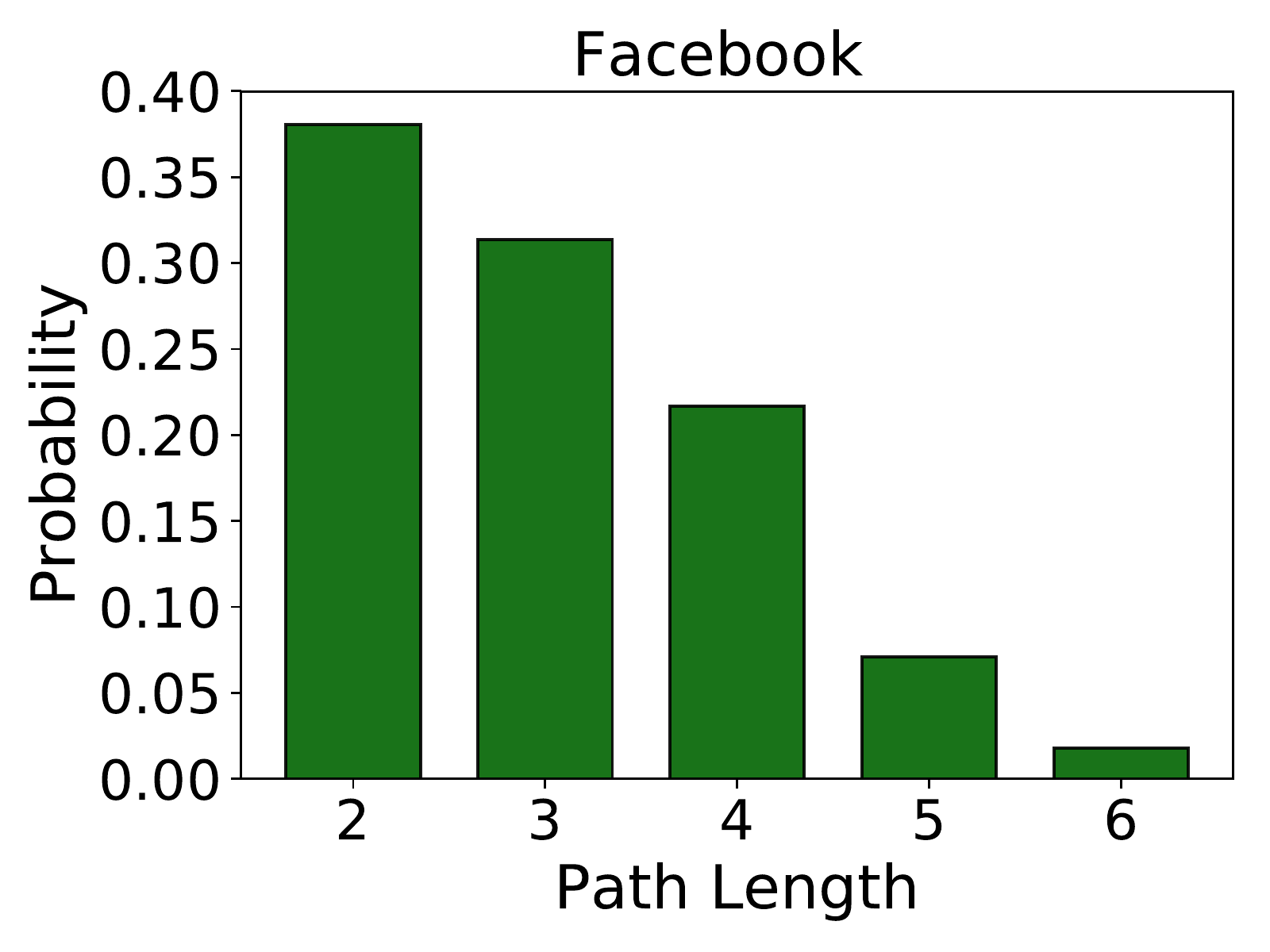}
}
\subfloat{
\includegraphics[width=.24\textwidth]{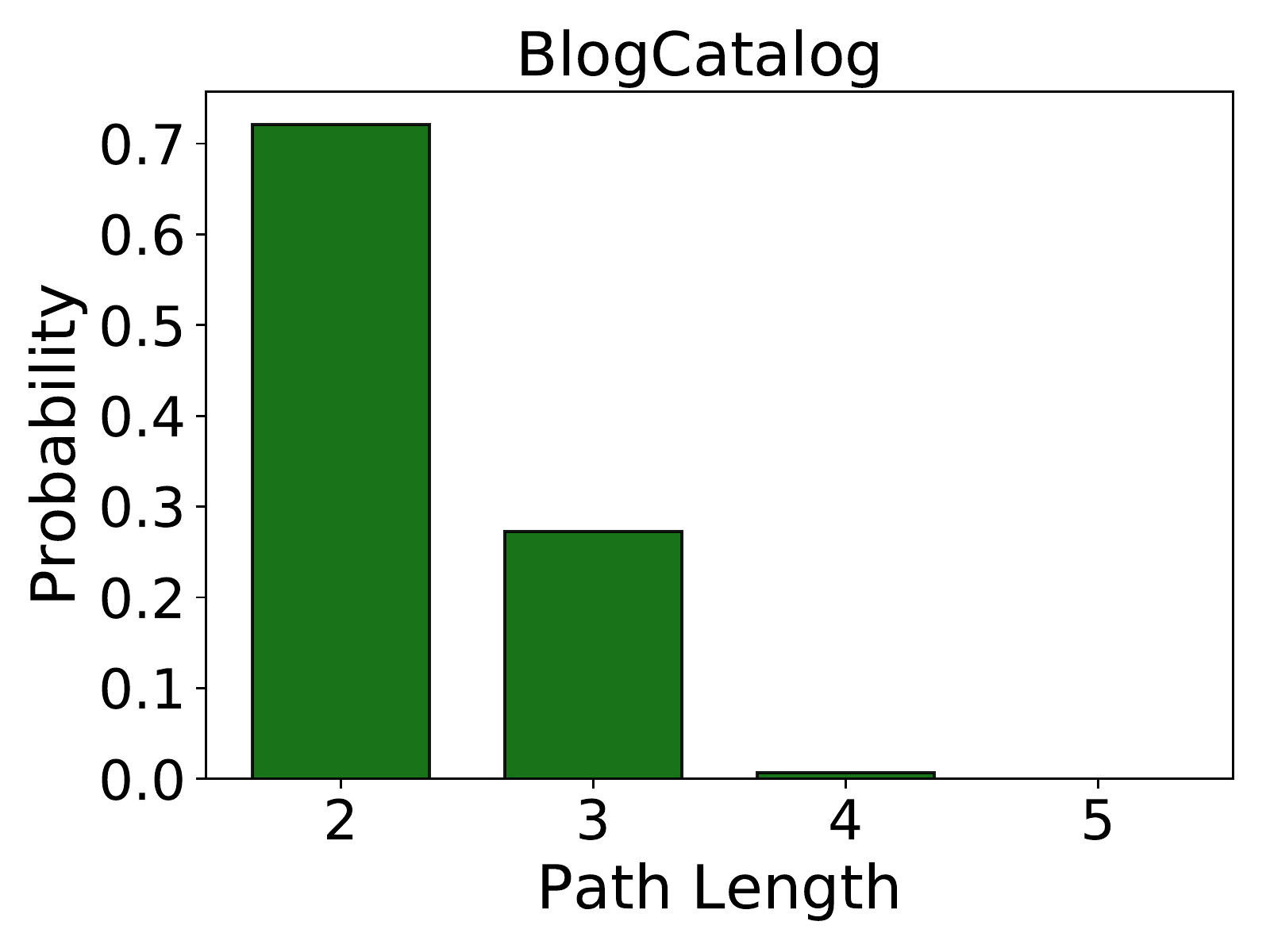}
}
\subfloat{
\includegraphics[width=.24\textwidth]{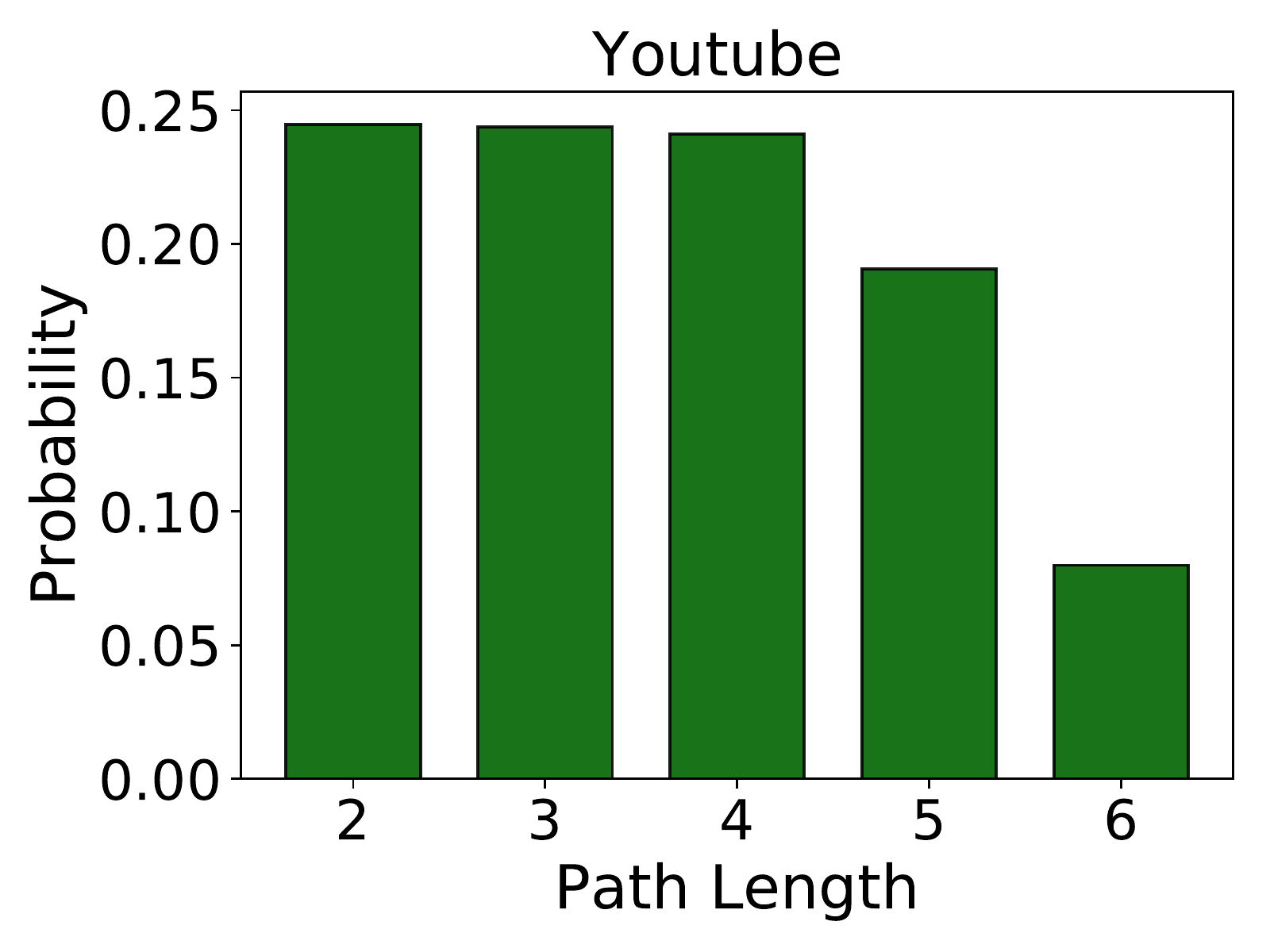}
}
\subfloat{
\includegraphics[width=.24\textwidth]{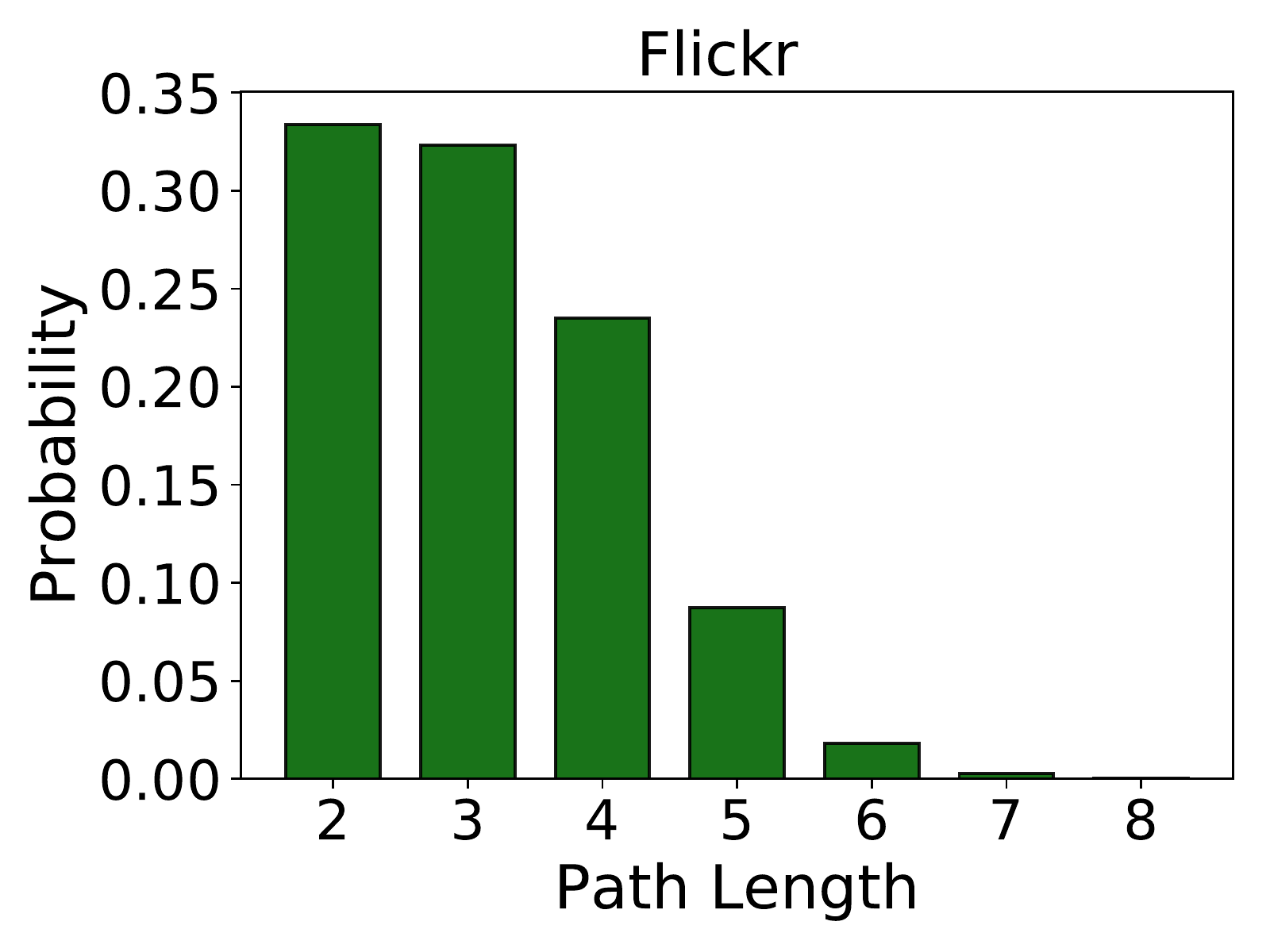}
}
\caption{Distributions of distances in the training set with downsampling for all datasets}
\label{fig:train}
\end{figure*}

\begin{table}[h!]
\caption{Statistics of Training  and Test data}

  \label{tab:train}
 
\begin{center}
\begin{tabular}{|l |c|c|c|}
  \hline
 \textbf{ Dataset}  &\textbf{ Nodes}  &\textbf{  Training pairs} & \textbf{ Test pairs} \Tstrut \\
 
  \hline
  
  Facebook    & 4,039 & 1,022,640 & 109,978 \Tstrut \\
  \hline
  BlogCatalog    & 10,312 & 1,409,700 & 88,316 \Tstrut \\
  \hline
  Youtube    & 1,134,890 & 2,452,757 & 184,413  \Tstrut \\
  \hline
  Flickr  & 1,715,255 & 2,579,437 & 112,967 \Tstrut  \\
  
  \hline
\end{tabular}
\end{center}

\end{table}

\subsection{Baseline}

As a baseline, we conduct an experiment where the  predictor is a simple linear regression. 
The input of the predictor is embeddings learnt by node2vec or Poincar\'{e} and the output is distance values between pair of nodes.
Since the state-of-the-art \cite{goyal2017 } observes the best performance with embedding dimensions $d = 128$   in primitive graph analysis tasks (e.g. link prediction, node classification), we set  $d = 128$
in our baseline experiment. However, we study the effect of dimension while we predict the distance using neural networks. Table \ref{tab:reg} shows MAE values for distance prediction besides the impact of different binary operations. We observe that node2vec achieves lower error values
than Poincar\'{e} specifically when we take the average of embedding vectors as the input of the predictor.

\begin{table}[htbp]
\caption{Mean Absolute Error (MAE) of shortest paths return by a Linear Regression }

\begin{center}
\begin{tabular}{|c|c|c|c|c|c|}

\hline
\textbf{Dataset}&
\textbf{Embedding}&
\multicolumn{4}{c|}{\textbf{MAE}} \Tstrut \\
  
\hline

  & &  $\ominus $ & $\oplus$ &$\oslash$ &$ \odot$  \Tstrut \\

  \multirow{2}{*}{Facebook}& node2vec &   0.679 &  0.663 &\textbf{ 0.546} & 0.653 \Tstrut \\ 
    
 & Poincar\'{e} &  0.801  & 0.788 & 0.656  & 0.767 \Tstrut \\ 
 
 \hline
 
 \multirow{2}{*}{BlogCatalog}& node2vec &  0.483 &  0.453 &  \textbf{0.407} & 0.423 \Tstrut \\ 
 & Poincar\'{e}&0.417    & 0.450  &0.436   & 0.447  \Tstrut \\  

\hline

 \multirow{2}{*}{Youtube}& node2vec &  0.708 &  0.722 &  \textbf{0.695}& 0.739\Tstrut \\ 
 & Poincar\'{e} & 0.983  & 1.267  & 1.195 & 1.099\Tstrut\\

\hline

  \multirow{2}{*}{Flickr }& node2vec  & 0.633  &  0.694 & \textbf{0.574} &  0.739 \Tstrut  \\ 
    
 & Poincar\'{e} & 0.944   & 0.891  & 0.831 & 0.822  \Tstrut  \\

\hline

\end{tabular}

\label{tab:reg}
\end{center}

\end{table}

\subsection{Results and Discussion}
\label{subsec:discu}
Using a feedforward neural network, we calculate the mean relative
and the mean absolute error during the test phase. The obtained error values for different embedding techniques, different embedding sizes and the different binary operators are
provided in Table~\ref{tab:err}. The best results are achieved by node2vec embeddings of size $128$ with an average MAE of $15\%$ across all datasets. The results indicate, that independent of the network size, our method achieves a quite low approximation error compared to the baseline.




\paragraph{Embedding Techniques and Embedding Size} The number of dimensions play an important role.
 Although
higher dimensions produce smaller errors, as the
dimension increases the time for generating embeddings and distance
approximation increases as well.

Reducing the size of embeddings in hierarchically organized networks was one goal underlying the development of Poincar\`e embeddings. Poincar\`e use a Hyperbolic space to alleviate overfitting and complexity issues
that Euclidean embeddings face especially if the data has
intrinsic hierarchical structure. Scale-free graphs in general and Social Networks in particular have been known to form innate
hierarchical structures~\cite{ravasz2003hierarchical}, which motivates the use of Poincar\`e embeddings. Moreover, we would  expect Poincar\`e embeddings to more accurately represent distances than node2vec embeddings in Euclidean space. However, Table~\ref{tab:err} shows that the error observed for node2vec embedding is remarkably lower than Poincar\`e embedding. 

The features learned in node2vec
are fundamentally tied to the random walk strategy. Node2vec adopts short random walks to
explore therefore it learns the structure of local neighborhoods. 
In scale-free networks,
the average shortest path distance grows logarithmically~\cite{barabasi2016network} , hence the shortest path distance stays as a local feature.
According to the result in Table~\ref{tab:err}, node2vec successfully learns distances specifically for embedding size $128$.

While, Poincar\`e  makes use of
hyperbolic spaces to encode both hierarchy and semantic similarity into a Poincar\`e ball. Regarding our results, Poincar\`e cannot capture features related to distances of nodes belong to different hierarchies. Table~\ref{tab:err} demonstrates that increasing dimensionality in Poincar\`e ball do not effect the accuracy of prediction in all datasets.
 
\begin{table*}[htbp]
\caption{Mean Absolute Error (MAE) and Mean Relative Error (MRE) of shortest paths utilizing different embedding techniques}

\begin{center}
\begin{tabular}{|c|c|c|c|c|c|c|c|c|c|c|c|  }
\hline
\textbf{Dataset}&
\textbf{Embedding}&
\textbf{Size}& 
\multicolumn{4}{c|}{\textbf{MAE}} &
\multicolumn{4}{c|}{\textbf{MRE}}  \Tstrut\\
  \hline  

  & & & $\ominus $ & $\oplus$ &$\oslash$ &$ \odot$&  $\ominus $ & $\oplus$ &$\oslash$ &$ \odot$ \Tstrut\\
   
  \multirow{4}{*}{Facebook}& \multirow{2}{*}{node2vec} & 32 & 0.480 & 0.415 & 0.233 &0.531 &0.175  & 0.164  &0.068  &0.188  \Tstrut\\ 
  
    & & 128 & 0.197  & 0.258  & \textbf{0.118} &0.217 & 0.071 & 0.099 & \textbf{0.038} &0.081 \Tstrut \\ 
    
 & \multirow{2}{*} {Poincar\'{e}} & 32 & 0.592 & 0.594 &  0.552&0.604  &0.214  & 0.211 & 0.218&0.212  \Tstrut \\
 
 & & 128 & 0.437 &  0.315 & 0.372 & 0.608 & 0.169 & 0.115  & 0.142 &0.246  \Tstrut\\ 

  \hline 

 \multirow{4}{*}{BlogCatalog}& \multirow{2}{*}{node2vec} & 32 & 0.277 & 0.242 & 0.197 & 0.193&0.092 & 0.103 & 0.067& 0.067  \Tstrut \\ 
 
    & & 128 & 0.220 & 0.275  &  0.159  & \textbf{0.154} &0.077& 0.119  & 0.064 &\textbf{0.059} \Tstrut \\ 
    
 & \multirow{2}{*} {Poincar\'{e}} & 32 & 0.338 & 0.338 & 0.343 & 0.338 &0.108 & 0.108 &0.112 &0.108  \Tstrut \\
 
 & & 128 & 0.331 &  0.354 &0.277  &0.338 &0.115 &  0.138 & 0.097 &0.108 \Tstrut \\ 
 
  \hline

 \multirow{4}{*}{Youtube}& \multirow{2}{*}{node2vec} & 32 & 0.676 & 0.265   & 0.455 & 0.625 &0.230 & 0.066  & 0.163 &0.223  \Tstrut\\ 
    & & 128 & 0.344  & \textbf{0.154 } & 0.174 &0.244 &  0.101& \textbf{0.034} &0.040 &0.061  \Tstrut \\
     
 & \multirow{2}{*} {Poincar\'{e}} & 32 & 1.095 & 0.708 & 1.134 & 0.774 &0.429 & 0.264 &0.446 &0.291 \Tstrut\\
 
 & & 128 & 1.270 & 1.185  & 1.746  & 0.771&  0.497 &  0.468 & 0.681 &0.262  \Tstrut\\ 
 
  \hline

 \multirow{4}{*}{Flickr}& \multirow{2}{*}{node2vec} & 32 & 0.699 & 0.295 & 0.564 & 0.525 &0.250 & 0.086  & 0.183 &0.198  \Tstrut\\ 
    & & 128 &0.238  & \textbf{0.168} & 0.181 & 0.222 &0.171 & \textbf{0.074 }& 0.178& 0.179 \Tstrut \\
     
 & \multirow{2}{*} {Poincar\'{e}} & 32 & 0.995 & 0.808 & 1.022 & 0.874 &0.349 & 0.284 &0.429 &0.278 \Tstrut\\
 
 & & 128 & 0.803 &  0.662& 0.807  &0.764& 0.397 & 0.432 & 0.566 & 0.364   \Tstrut\\ 
 
  \hline
 
\end{tabular}
\label{tab:err}
\end{center}

\end{table*}

\begin{figure*}[htbp]  
\centering

\subfloat{
\includegraphics[width=.24\textwidth]{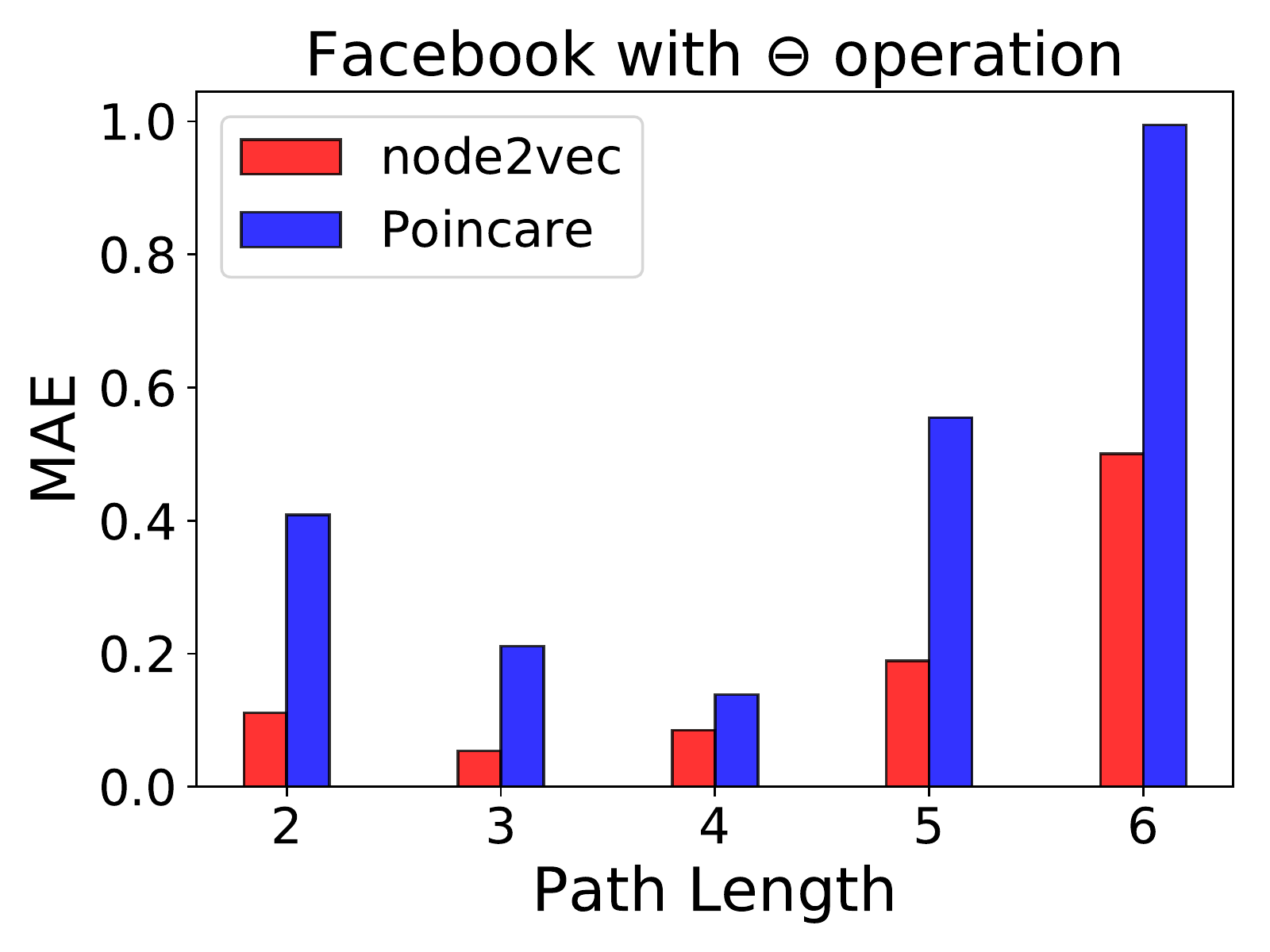}
}
\subfloat{
\includegraphics[width=.24\textwidth]{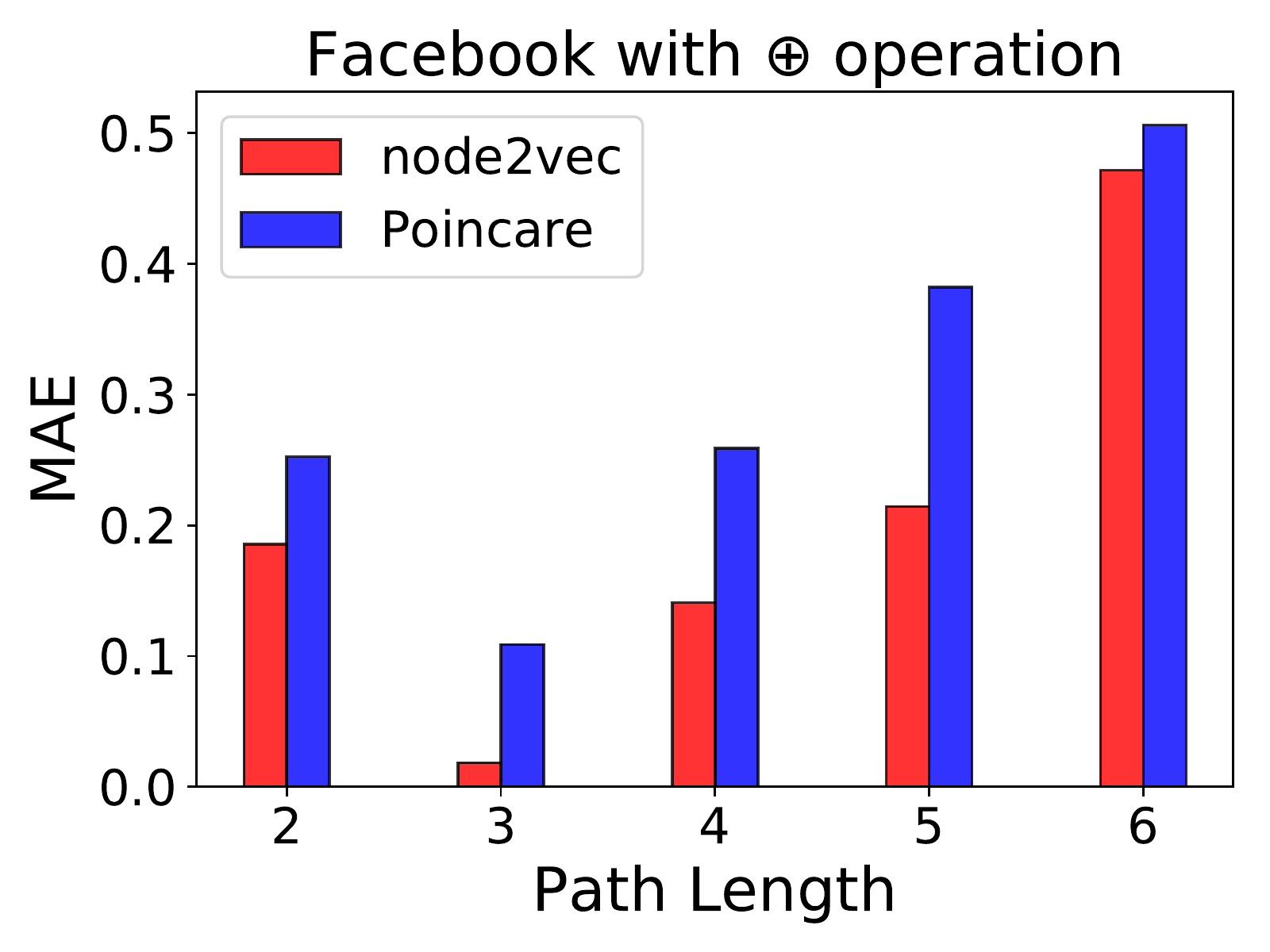}
}
\subfloat{
\includegraphics[width=.24\textwidth]{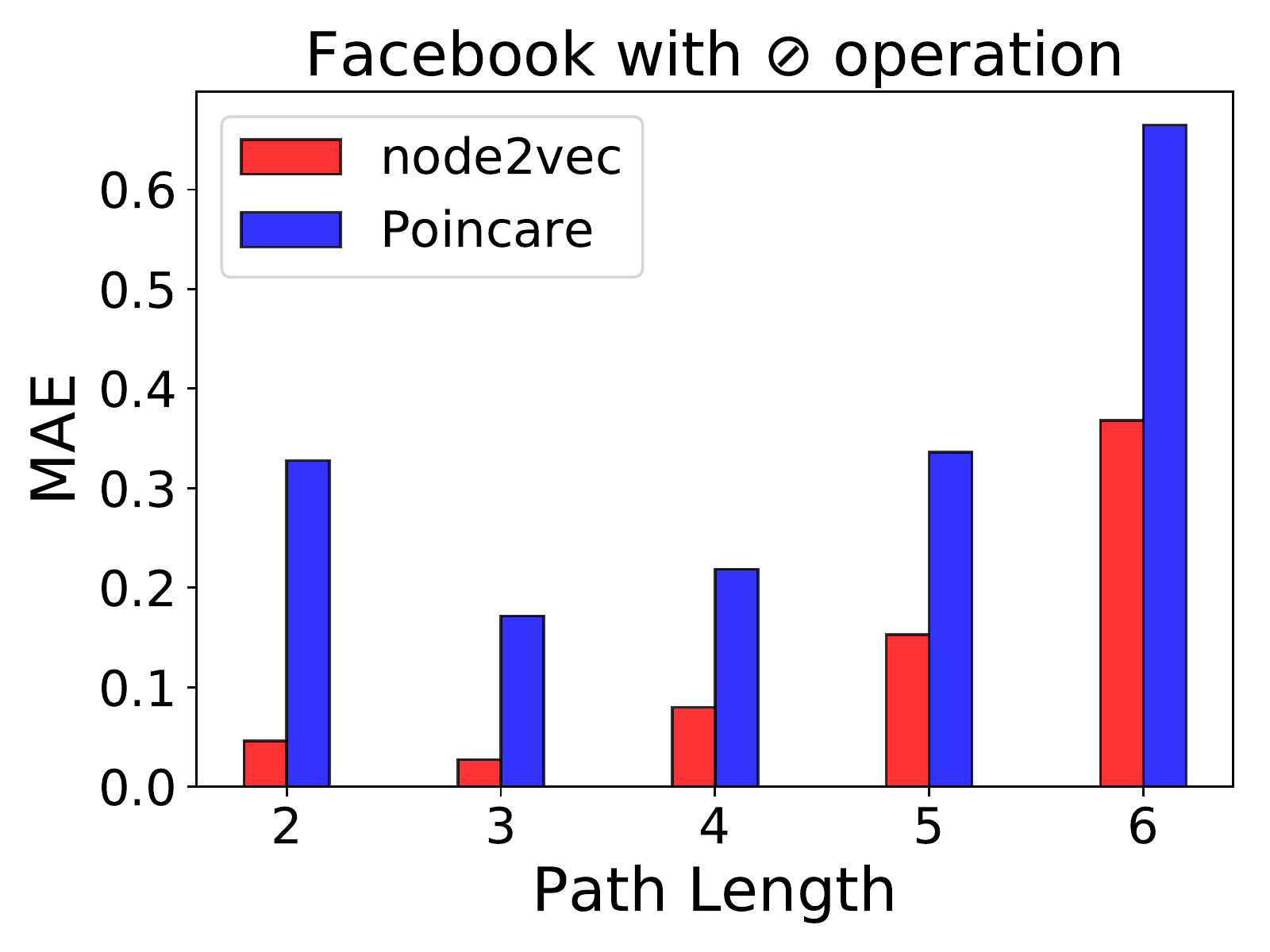}
}
\subfloat{
\includegraphics[width=.24\textwidth]{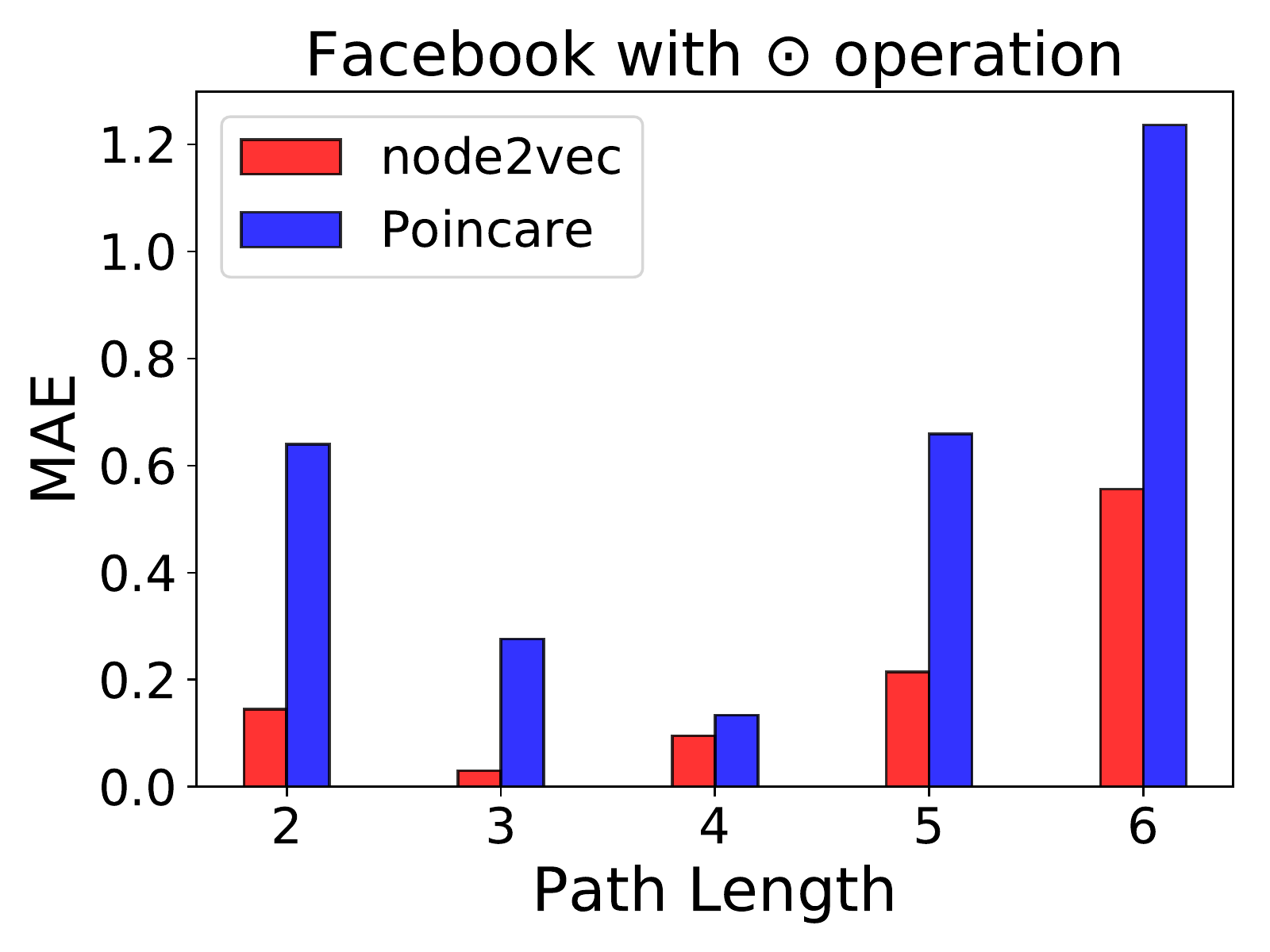}
}

\subfloat{
\includegraphics[width=.24\textwidth]{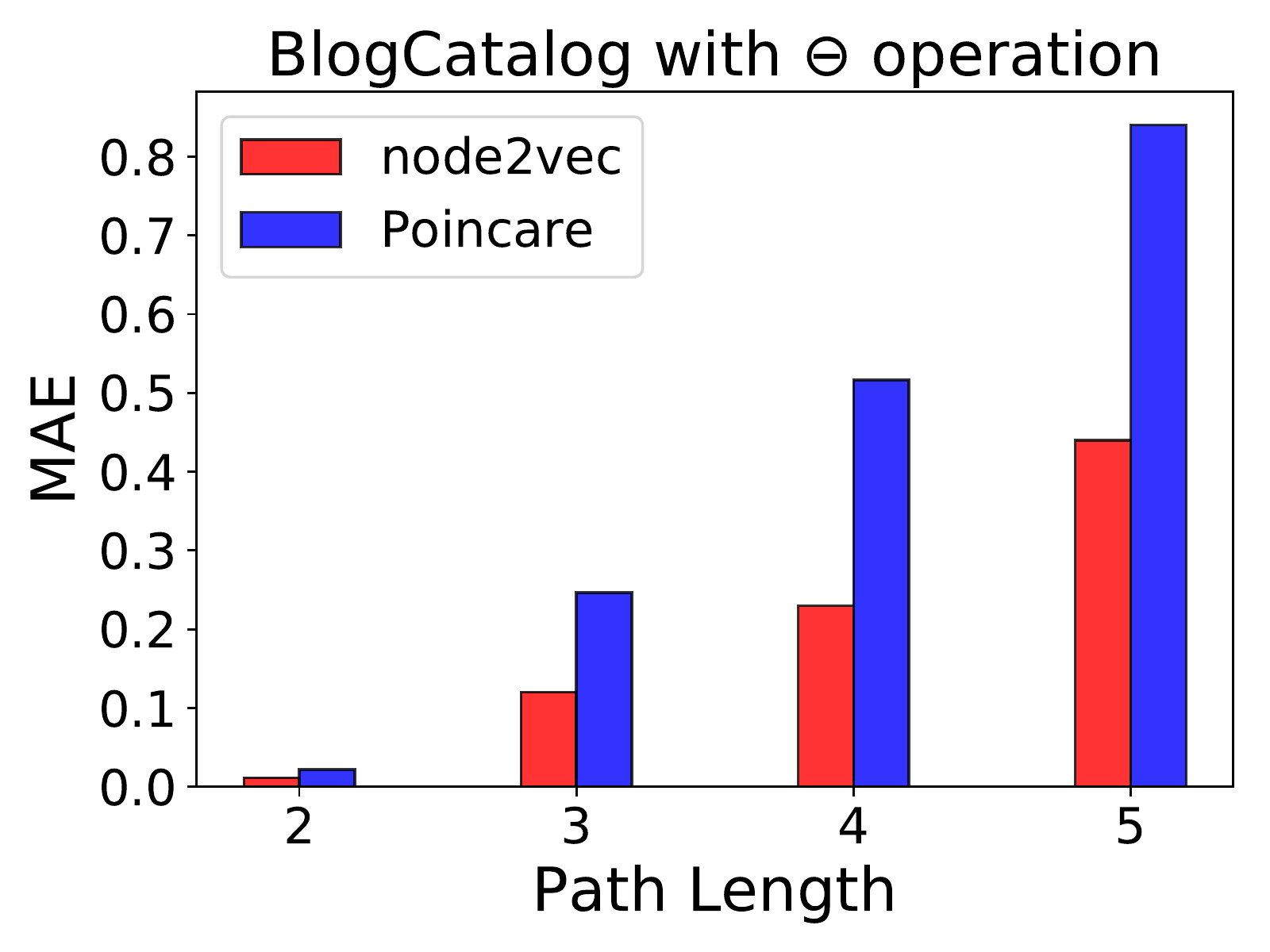}
}
\subfloat{
\includegraphics[width=.24\textwidth]{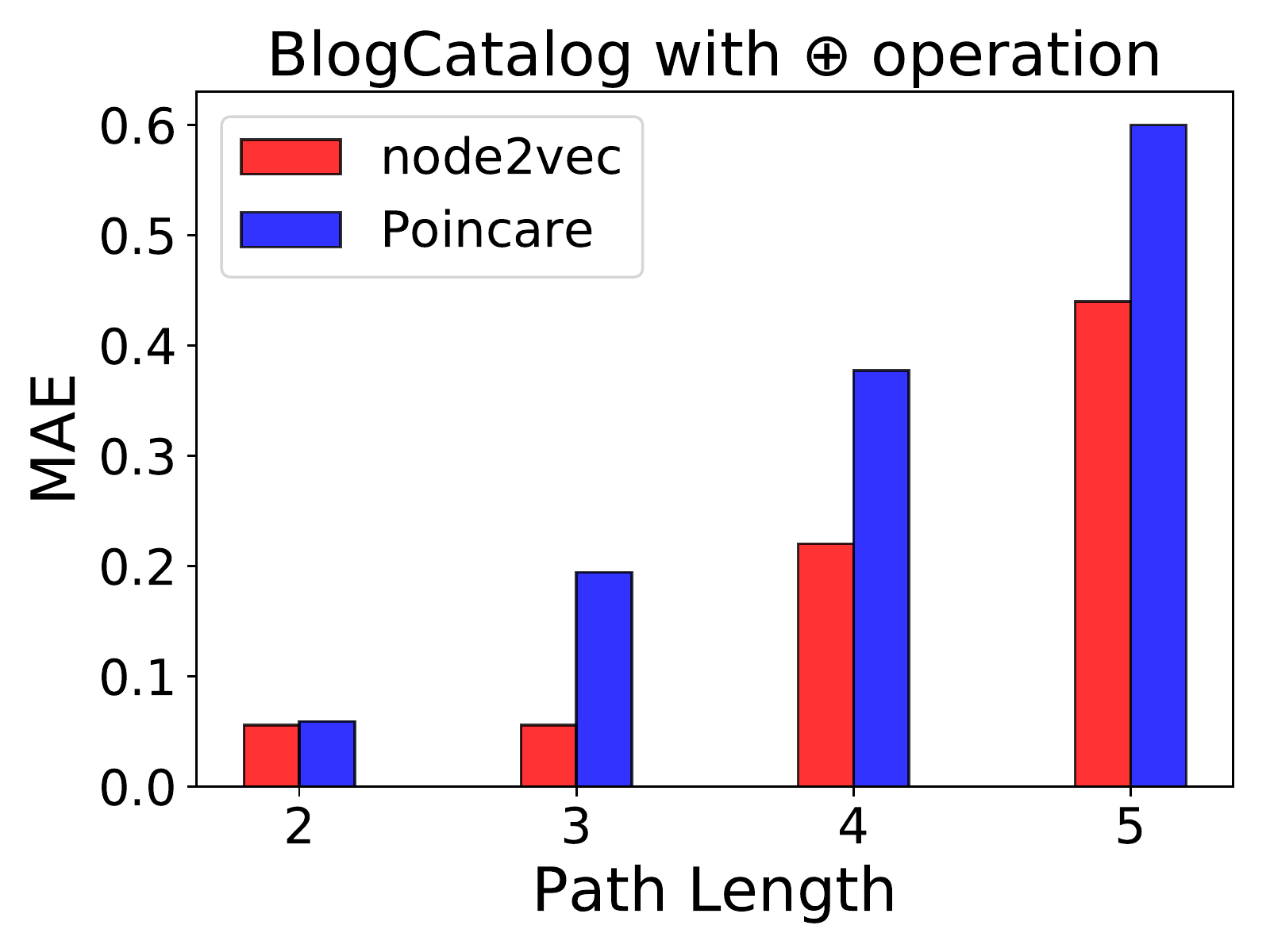}
}
\subfloat{
\includegraphics[width=.24\textwidth]{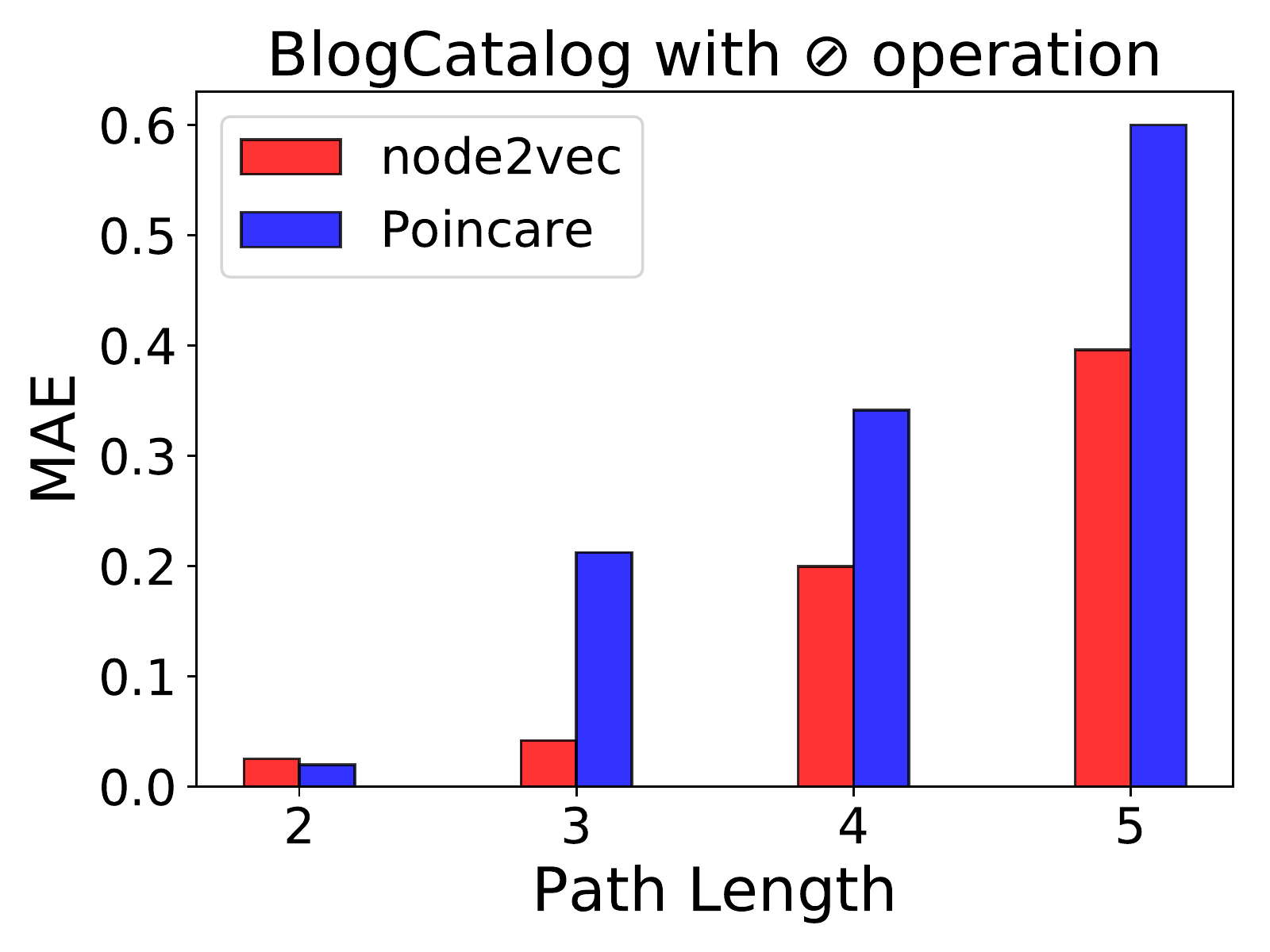}
}
\subfloat{
\includegraphics[width=.24\textwidth]{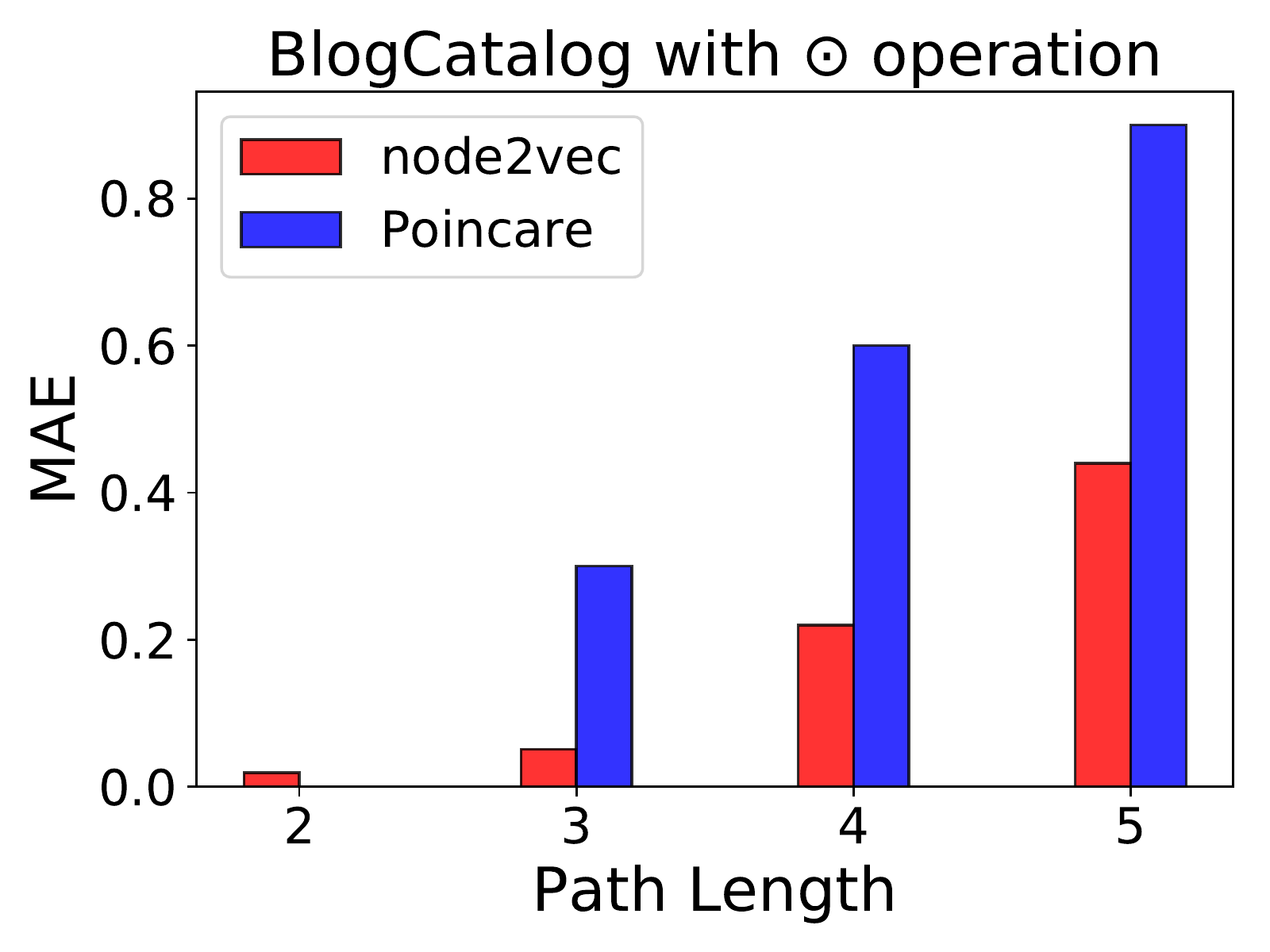}
}

\subfloat{
\includegraphics[width=.24\textwidth]{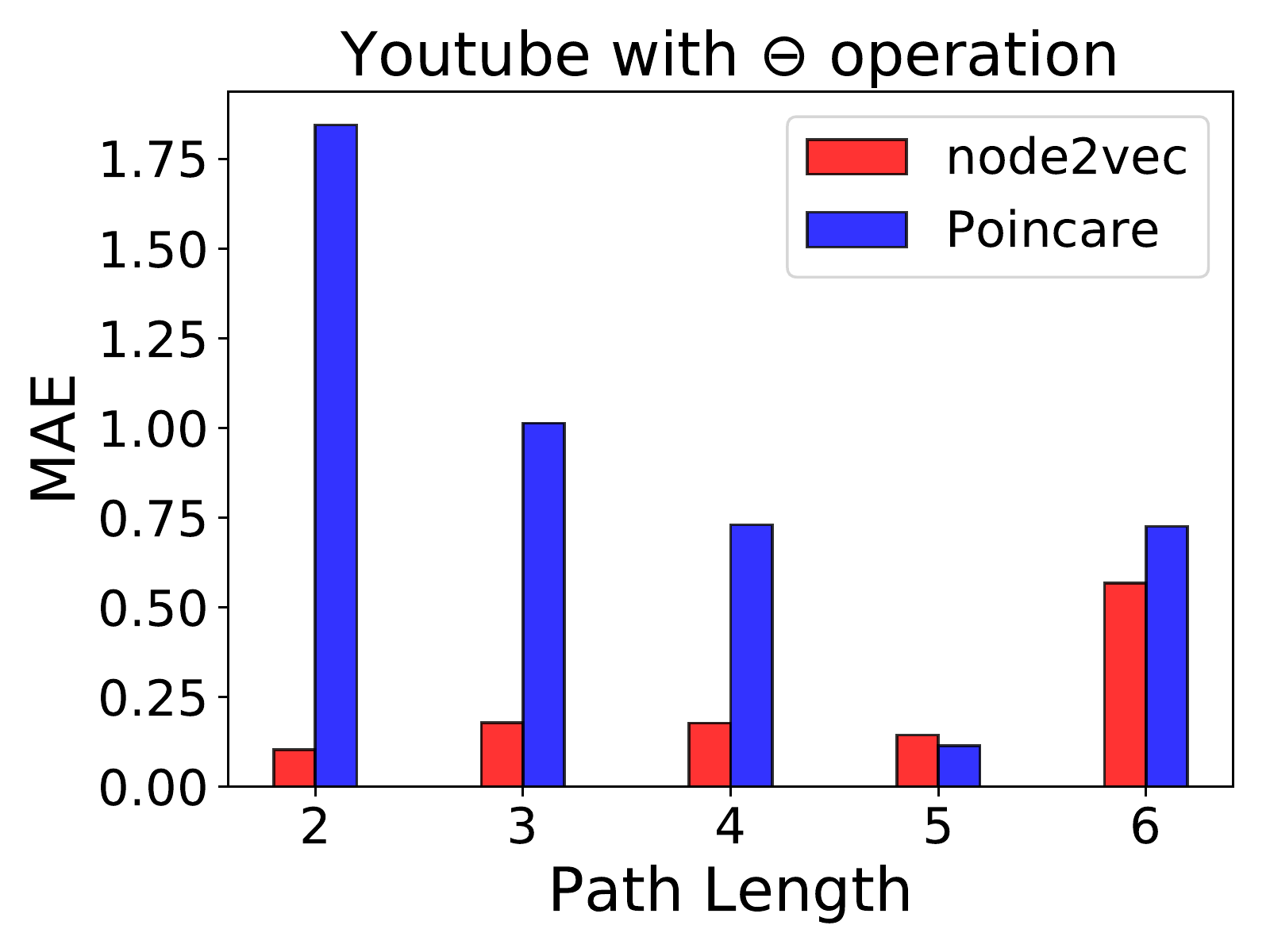}
}
\subfloat{
\includegraphics[width=.24\textwidth]{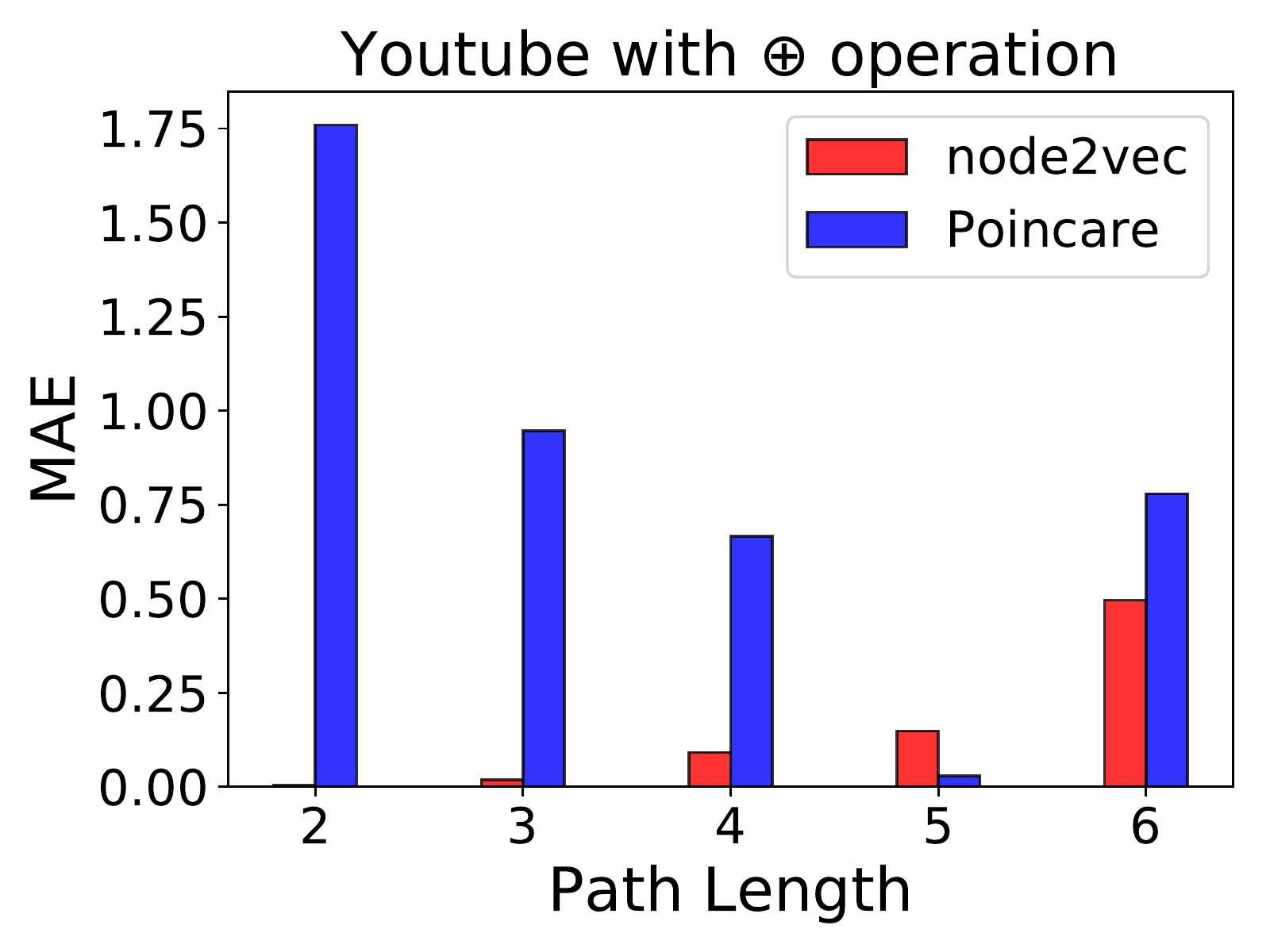}
}
\subfloat{
\includegraphics[width=.24\textwidth]{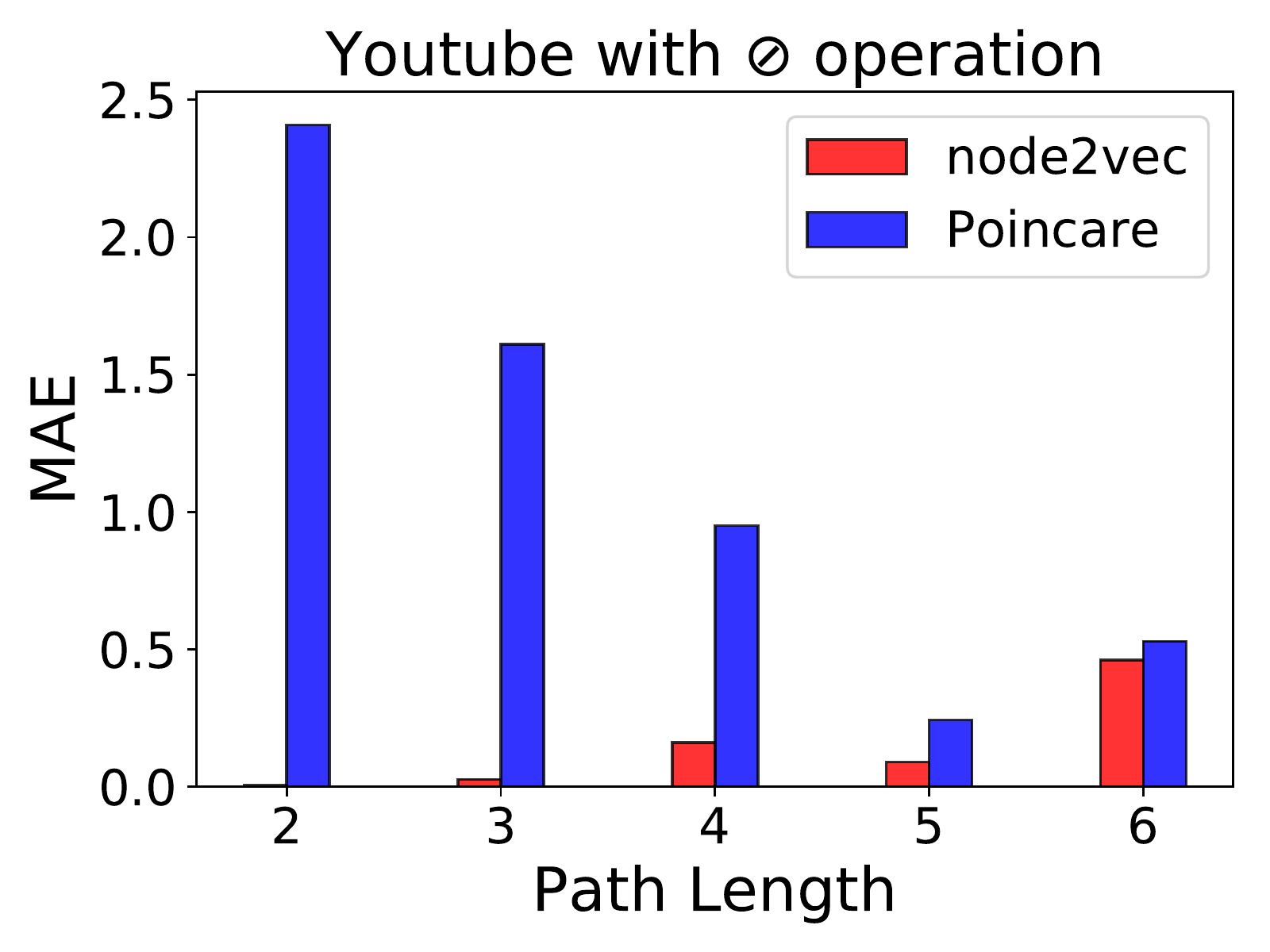}
}
\subfloat{
\includegraphics[width=.24\textwidth]{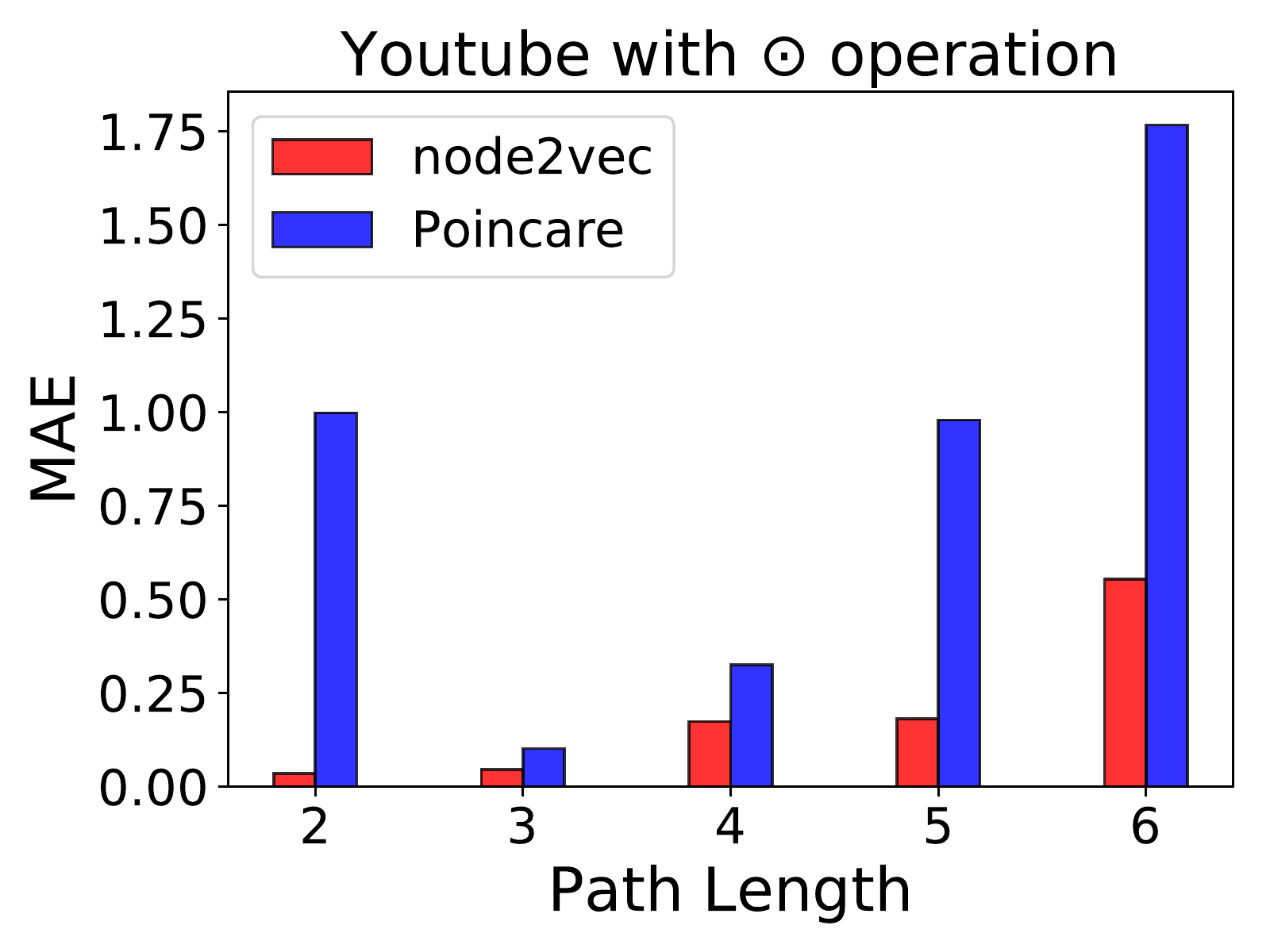}
}

\caption{Mean Absolute Error (MAE) of different path
lengths comparing node2vec and Poincar\'{e} embeddings with size $128$ for Facebook, BlogCatalog and Youtube datasets.}

\label{fig:error}
\end{figure*}

\paragraph{Error Distribution over Path Lengths} We explore the accuracy of
predictions for paths of different lengths. 
Figure~\ref{fig:error} shows the
mean absolute errors per path length on three
graphs. Error values are estimated for different binary operations on embeddings of size $128$. Observe that the larger errors caused by longer paths utilizing node2vec embeddings. In one hand, we do not have enough samples for longer distances in the training set. On the other hand, node2vec fails to learn structural features of faraway nodes. Similarly, long paths cause high errors using Poincar\`e embedings in all three datasets. It can be the effect of locality property of the Poincar\'{e} distance which places leaf nodes to the boundary. Therefore, the distance between leaf nodes of different hierarchies cannot be preserved according the original graph neighborhood.

\paragraph{Effect of Binary Operators} 
To generate feature representations of training pairs, we compose the learned embeddings of the individual nodes using simple binary operators. The binary operators are listed in Table~\ref{tab:opr}.
This compositionality lends node2vec and Poincar\'{e} to the prediction task
involving nodes. As Table~\ref{tab:err} shows, binary operators do not have a consistent behavior over different datasets and different dimension sizes. For instance, the average operator outperforms others in Facebook graph while concatenation works better for Youtube dataset.
As a future work, we would like to explore
the reasons behind the unstable behavior of operators as well as effect of some other operators on prediction.

\begin{figure*}[htbp]
\centering

\subfloat{
\includegraphics[width=.24\textwidth]{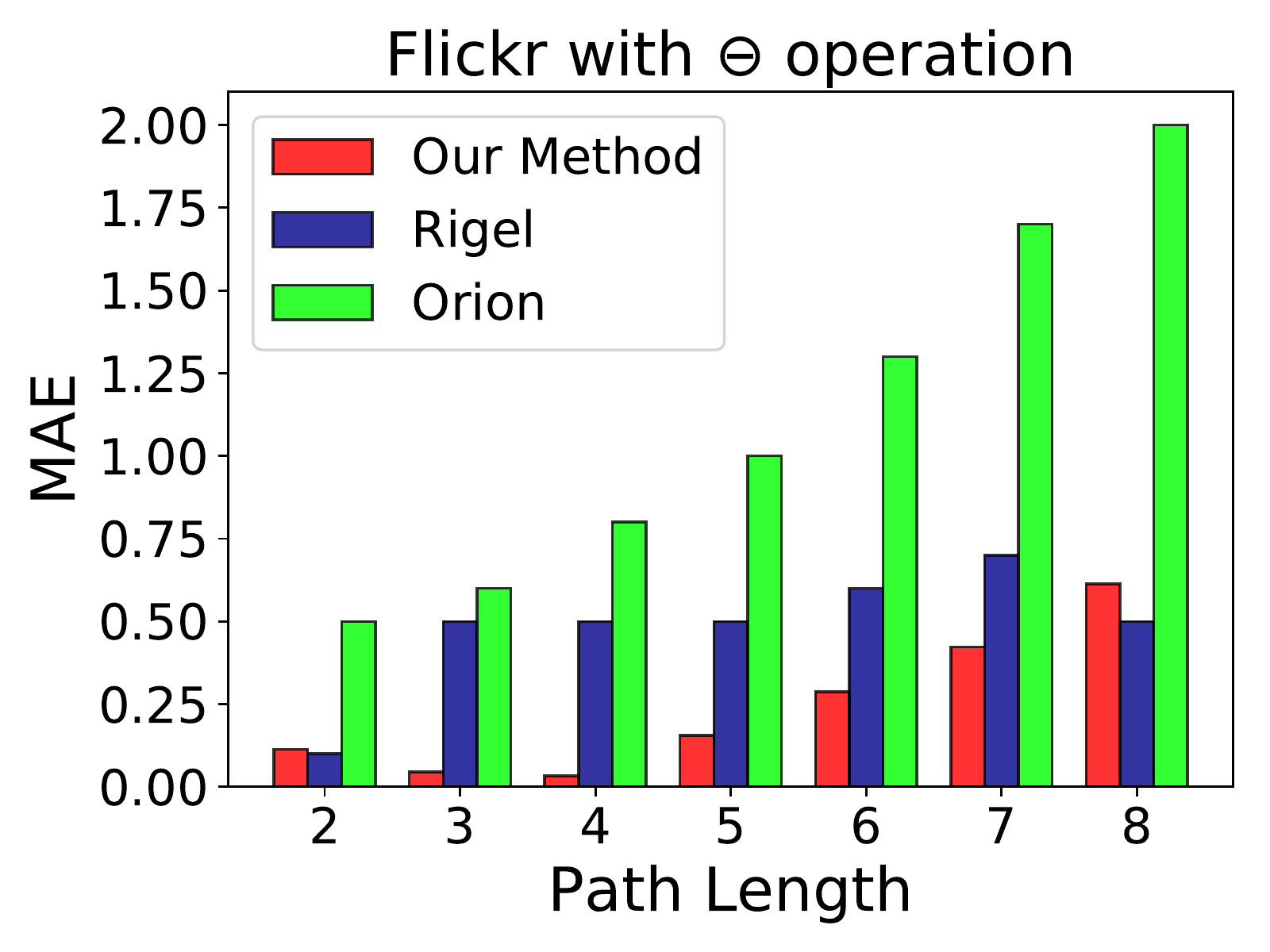}
}
\subfloat{
\includegraphics[width=.24\textwidth]{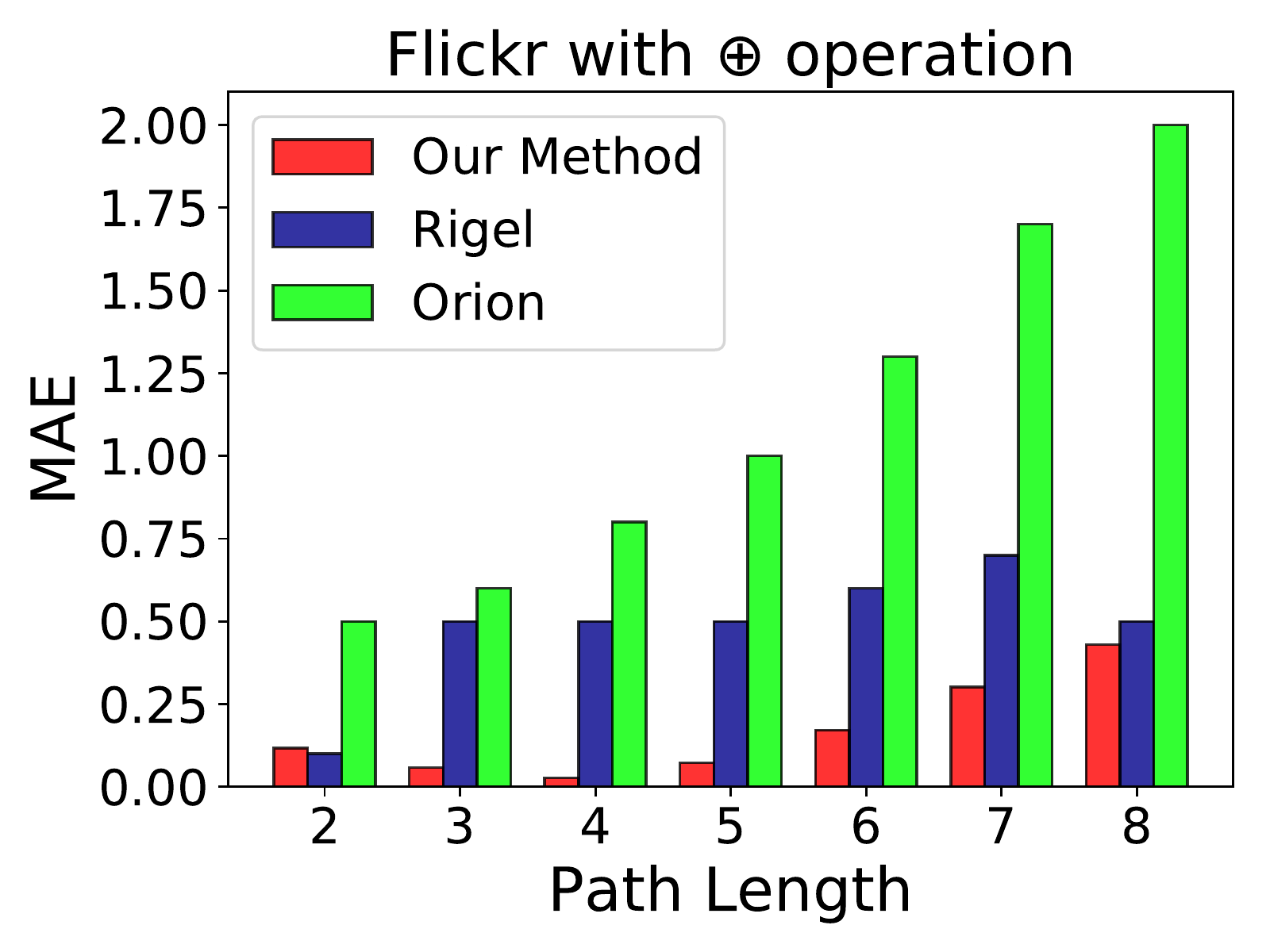}
}
\subfloat{
\includegraphics[width=.24\textwidth]{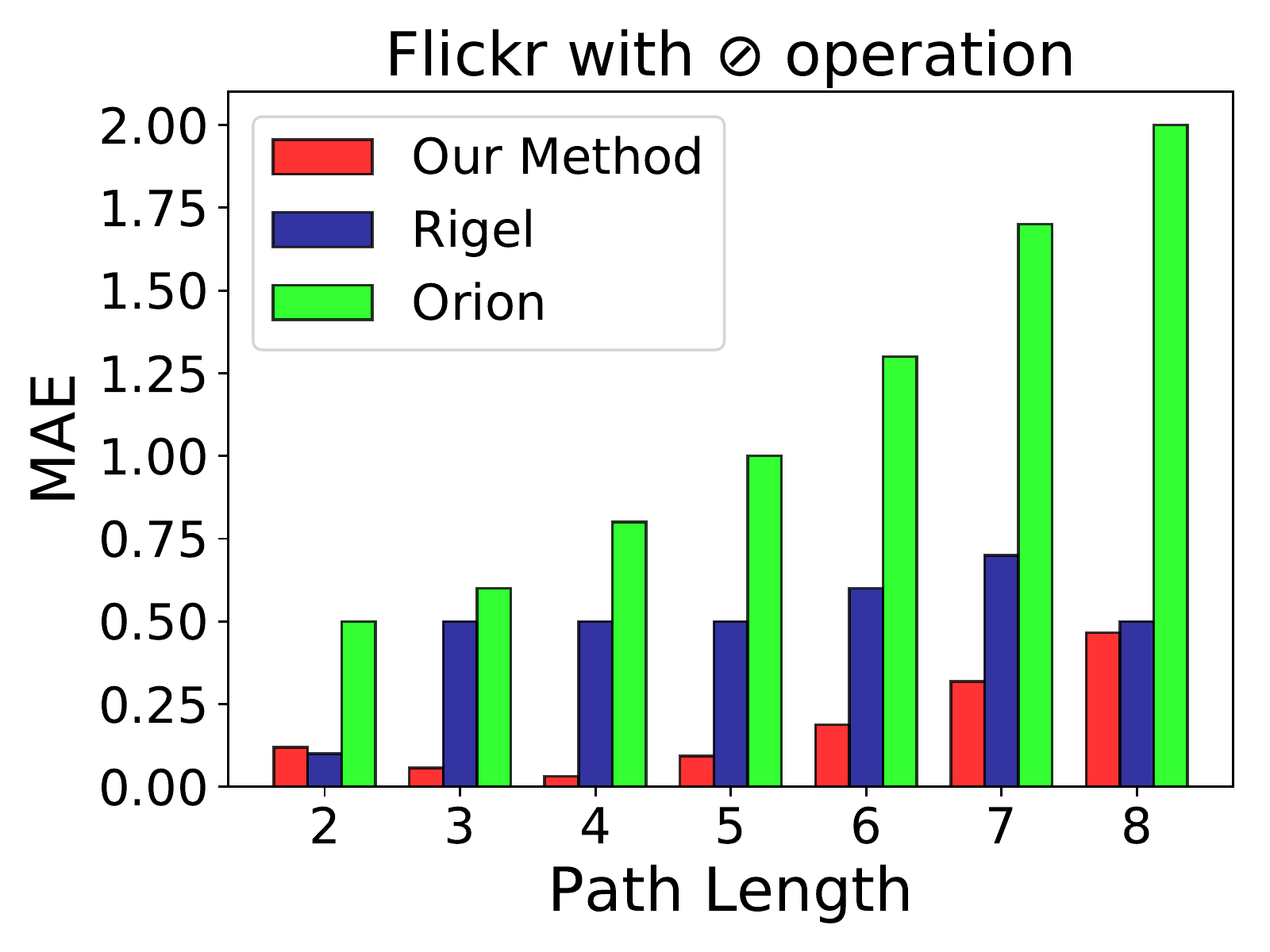}
}
\subfloat{
\includegraphics[width=.24\textwidth]{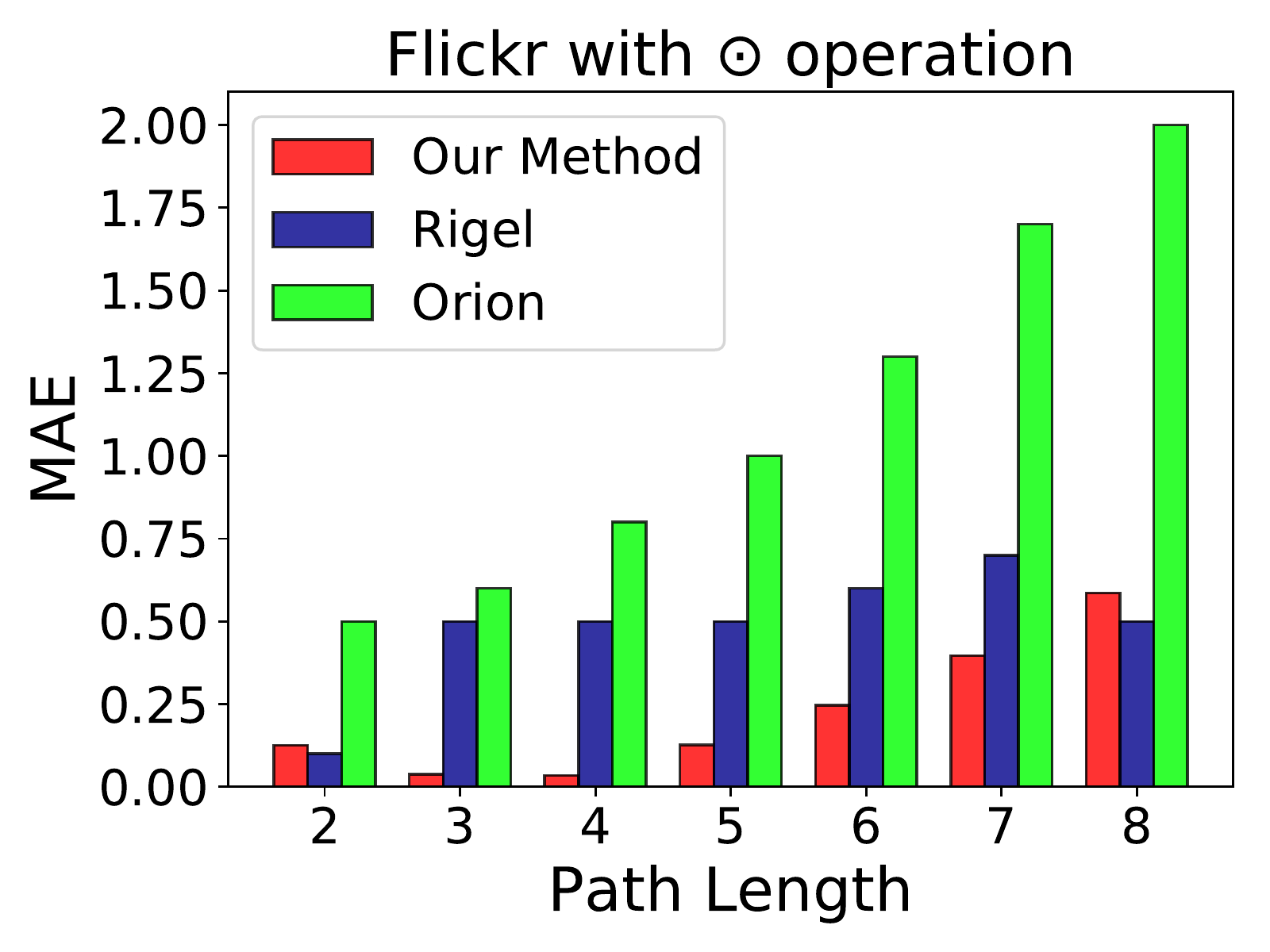}
}

\caption{Mean Absolute Error (MAE) of shortest paths returned by our method, Orion and Rigel on the Flickr dataset. }
\label{fig:compare}
\end{figure*}

\paragraph{Comparison to the State-of-the-Art} In Figure~\ref{fig:compare}, we plot the MAE for different path lengths using our method against the two state-of-the-art methods Rigel~\cite{ref9}
and Orion~\cite{ref8} on the Flickr dataset. Errors are calculated when the input of the predictor is node2vec embeddings of size $128$. We can observe that Orion shows the highest MAE value for all paths.
In general, our method consistently and significantly outperforms Rigel specifically for shorter paths.  

\section{Conclusion and future work}
\label{sec:con}

Traditional methods for computing shortest path distances no
longer scale to today's massive graphs with millions of nodes
and billions of edges. 
Motivated by landmarked-based approaches, we propose a new method
that approximates node distances by first embedding graphs
into an embedding space. We utilized two recent graph embedding techniques and fed their vectors into a feedforward neural network. The results are impressive. Our method
produces shortest distances for the large majority of node
pairs, matching the most accurate of ground truth. And it does
this quickly, returning results in a linear time. 

We plan to handle longer paths in larger graphs,
where we face the higher errors.
One way is to apply other embedding techniques such as HARP~\cite{chen2018} which learns the  structure of neighborhoods using graph coarsening.
Another way is to reach a balance training set and to use other neural networks such as Siamese neural networks~\cite{koch2015siamese}.

\section*{Acknowledgment}

The presented work was developed within the Provenance Analytics project funded by the German Federal Ministry of Education and Research, grant agreement number 03PSIPT5C.

\bibliographystyle{IEEEtran}
\bibliography{sample-bibliography}

\end{document}